\documentclass[journal]{IEEEtran}
\newcommand{\qed}{\nobreak \ifvmode \relax \else
      \ifdim\lastskip<1.5em \hskip-\lastskip
      \hskip1.5em plus0em minus0.5em \fi \nobreak
      \vrule height0.75em width0.5em depth0.25em\fi}
			
\usepackage{algorithm}
\usepackage{algorithmic}			
\usepackage{array,booktabs,tabularx}			

\ifCLASSINFOpdf
  \usepackage[pdftex]{graphicx}
  \graphicspath{{./}{Figs/}}
  \DeclareGraphicsExtensions{.pdf,.jpeg,.png}
\else
\fi
\usepackage{cite}
\usepackage{amssymb}
\usepackage{amsmath}
\usepackage{mathtools}
\usepackage{bigdelim}
\usepackage{tikz}
\usepackage{bbm}
\usepackage{mathtools}
\usepackage{centernot}

\usepackage{enumerate}
\usepackage{subcaption}
\usepackage{pgfplots}
\usepackage[caption=false,font=footnotesize]{subfig}
\usepackage{undertilde}
\usepackage{verbatim}
\usetikzlibrary{arrows,calc}
\usepackage{relsize}

\usepackage{booktabs}

\begin{document}
\title{Personalized Risk Scoring for Critical Care Prognosis using Mixtures of Gaussian Processes}

\author{Ahmed~M.~Alaa,~\IEEEmembership{Member,~IEEE}, Jinsung~Yoon, Scott~Hu, {\it MD}, and~Mihaela~van~der~Schaar,~\IEEEmembership{Fellow,~IEEE}
\thanks{A. Alaa, J. Yoon and M. van der Schaar are with the Department of Electrical Engineering, University of California, Los Angeles (UCLA), CA, 90095, USA (e-mail: ahmedmalaa@ucla.edu, jsyoon0823@ucla.edu, mihaela@ee.ucla.edu).} 
\thanks{S. Hu is with the Division of Pulmonary and Critical Care Medicine, Department of Medicine, David Geffen School of Medicine, University of California, Los Angeles (UCLA), CA, 90095, USA (email: scotthu@mednet.ucla.edu).} 
}
\markboth{XXXX, ~Vol.~XX, No.~X, XXXX~2016}%
{Alaa \MakeLowercase{\textit{et al.}}: Personalized Risk Scoring for Critical Care Prognosis using Mixtures of Gaussian Processes}
\maketitle
\begin{abstract} 
\textit{Objective}: In this paper, we develop a personalized real-time risk scoring algorithm that provides timely and granular assessments for the clinical acuity of ward patients based on their (temporal) lab tests and vital signs; the proposed risk scoring system ensures timely intensive care unit (ICU) admissions for clinically deteriorating patients. \textit{Methods}: The risk scoring system learns a set of latent patient {\it subtypes} from the offline electronic health record data, and trains a mixture of {\it Gaussian Process (GP) experts}, where each expert models the physiological data streams associated with a specific patient subtype. Transfer learning techniques are used to learn the relationship between a patient's latent subtype and her static admission information (e.g. age, gender, transfer status, ICD-9 codes, etc). \textit{Results}: Experiments conducted on data from a heterogeneous cohort of 6,321 patients admitted to Ronald Reagan UCLA medical center show that our risk score significantly and consistently outperforms the currently deployed risk scores, such as the Rothman index, MEWS, APACHE and SOFA scores, in terms of timeliness, true positive rate (TPR), and positive predictive value (PPV). \textit{Conclusion}: Our results reflect the importance of adopting the concepts of personalized medicine in critical care settings; significant accuracy and timeliness gains can be achieved by accounting for the patients' heterogeneity. \textit{Significance}: The proposed risk scoring methodology can confer huge clinical and social benefits on more than 200,000 critically ill inpatient who exhibit cardiac arrests in the US every year.  
\end{abstract} 
\IEEEpeerreviewmaketitle{}
\section{Introduction}
\IEEEPARstart{C}{ritically} ill patients who are hospitalized in regular wards with solid tumors, hematological malignancies, neutropenia, or those who are recipients of stem cell (or bone marrow) transplants, or upper-gastrointestinal surgeries, are vulnerable to a wide range of adverse outcomes, including post-operative complications \cite{churpek2014using, kirkland2013clinical, rothman2013development, clifton2012gaussian, prytherch2010views, young2003inpatient, finlay2014measuring, pimentel2013modelling}, cardiopulmonary arrest \cite{kause2004comparison, hogan2012preventable}, and acute respiratory failure \cite{mokart2013delayed}. All these adverse events can lead to an unplanned ICU transfer \cite{kirkland2013clinical}, the timing of which plays a major role in determining clinical outcomes, since the efficacy of acute care interventions (including thrombolytic agents, aspirin and $\beta$-blockers, mechanical ventilation, etc) depends substantially on the timeliness of their application. Recent medical studies have confirmed that delayed transfer to the ICU is strongly correlated with mortality and morbidity \cite{Datades1,mokart2013delayed,young2003inpatient}, and according to the {\it Joint Commission}\footnote{A nonprofit organization that accredits hospitals and gathers data related to adverse events.}, around 29$\%$ of (narcotic-related) bedside adverse events reported during the period from 2004 to 2011 were resulting from improper post-operative (or pre-operative) monitoring of patients\cite{jointcomm2014}. \\ 

\begin{figure}[t!]
    \centering
    \includegraphics[width=3.5 in]{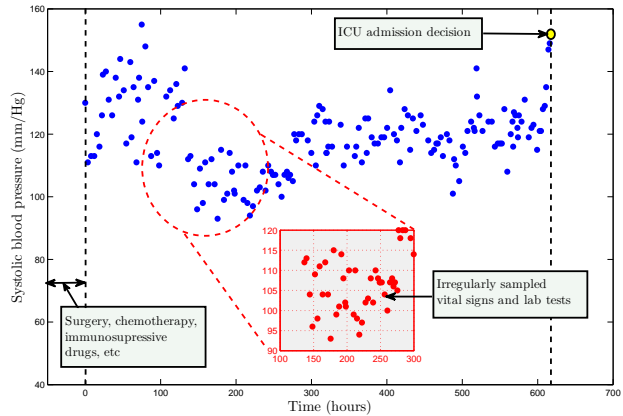}     
    \caption{An exemplary physiological stream for a patient hospitalized in a regular ward.}
		\label{prevwfig}
\end{figure}

In the light of the above, the {\it Institute for Healthcare Improvement}\footnote{A nonprofit organization focused on motivating and building the will for change, partnering with patients and health care professionals to test new models of care.} recommended implementing {\it rapid response teams} that could provide rapid bedside treatments for deteriorating patients in order to decrease hospital mortality rates and avoid serious events \cite{Datades1}. Other civil organizations, such as {\it LeahsLegacy}\footnote{Refer to the (Wall Street Journal) article in \cite{WSJ}.}, are advocating continuous electronic monitoring of patients on narcotics in the hospital. Improved critical care monitoring and prognosis for patients in wards is expected to have a significant clinical and social impact. For instance, qualitative studies (based on experts' opinions) have suggested that up to 50$\%$ of cardiopulmonary arrests on general (medical or surgical) wards could be prevented by earlier transfer to the ICU \cite{hershey1982outcome, franklin1994developing, schein1990clinical}. Since over 200,000 in-hospital cardiac arrests occur in the U.S. each year \cite{merchant2011incidence}, improved patient monitoring and vigilant care in wards would translate to a large number of lives saved yearly. \\

In an attempt to address the challenges above, hospitals have been investigating and investing in prognostic risk scoring systems that quantify and anticipate the acuity of critically ill inpatients in real-time based on their (temporally evolving) physiological signals in order to ensure timely ICU transfer \cite{churpek2014using, kirkland2013clinical, rothman2013development, clifton2012gaussian, prytherch2010views, young2003inpatient, finlay2014measuring}. Prognosis in hospital wards is feasible since unanticipated adverse events are often preceded by disorders in a patient's physiological parameters \cite{kause2004comparison, hogan2012preventable}. However, the subtlety of evidence for clinical deterioration in the physiological parameters makes the problem of constructing an ``informative" risk score quite challenging: overestimating a patient's risk can lead to alarm fatigue and inefficient utilization of clinical resources \cite{subbe2001validation}, whereas underestimating her risk can undermine the effectiveness of consequent therapeutic interventions \cite{liu2012adverse,mokart2013delayed}. \\

Recent systematic reviews have shown that currently deployed expert-based risk scores, such as the MEWS score \cite{morgan1997early}, provide only modest contributions to clinical outcomes \cite{tsien1997poor,cvach2012monitor, bliss2000behavioural}. Alternatives for expert-based risk scores can be constructed by training a risk scoring model using the data available in the electronic health records (EHR) \cite{kirkland2013clinical}. Recently, a data-driven risk score, named the Rothman index, has been developed using regression analysis \cite{rothman2013development}, and was shown to outperform the MEWS score and its variants \cite{finlay2014measuring}. However, this score lacks a principled model for the hospitalized patient's physiological parameters, and is mainly constructed using a ``one-size-fits-all" approach that leaves no room for personalized risk assessment that is tailored to the individual patient (see Subsection \ref{sub1.3} for more details). Personalized models that account for the patient's individual traits are anticipated to provide significant accuracy and granularity in risk assessments \cite{snyderman2012personalized}. The goal of this paper is to develop a principled and personalized risk scoring methodology that overcomes the limitations of the current state-of-the-art risk scores, and boosts the quality of care in regular hospital wards. Contributions are summarized in the next subsection. 

\begin{table*}[t]
  \centering
  \begin{tabular}{p{1.75cm}|p{2.5cm}|p{5.5cm}|p{5.75cm}}
    \hline
		\hline
    Reference & Risk scores & Details & Limitations \\
    \hline
		 &  &  & \\
    \cite{prytherch2010views, alam2014impact, henry2015targeted, goldhill2005physiologically, parshuram2009development, subbe2001validation, morgan1997early} & MEWS, ViEWS and TREWS & Expert-based risk assessment methodologies (also known as ``track and trigger" systems) & \begin{itemize} \item Neither personalized nor data-driven, does not take advantage of the EHR. \item Modest performance reported by recent systematic reviews in \cite{tsien1997poor, cvach2012monitor, bliss2000behavioural}. \end{itemize}  \\
		&  &  & \\
		\cite{vincent1996sofa, jones2009sequential, ferreira2001serial, yu2014comparison} & SOFA & A combination of organ dysfunction scores for respiratory, coagulation, liver, cardiovascular and renal systems. Originally developed for predicting mortality in ICU patients, but was shown in \cite{yu2014comparison} to function as a prognostication tool for non-ICU ward patients. &  \begin{itemize} \item Not personalized, i.e. uses the same scoring scheme for all patients (see Table 3. in \cite{vincent1996sofa}). \item Does not consider correlations between organ dysfunction scores and endpoint outcomes. \item Predictions can corporate the mean statistics of the computed score over time but does not consider the full temporal trajectory. \end{itemize}    \\
		&  &  & \\
		\cite{knaus1985apache, goel2003apache, yu2014comparison} & APACHE II and III &  A disease severity score used for ICU patients (usually applied within 24 hours of admission of a patient to the ICU \cite{knaus1985apache}). It has been shown in \cite{yu2014comparison} that it can be used for prognostication in regular wards.  & \begin{itemize} \item Does not consider the temporal trajectory of score evaluations during the patients stay in ICU (or in the ward). \end{itemize} \\
		&  &  & \\
		\cite{finlay2014measuring, rothman2013development} & Rothman index & A regression-based data-driven model that utilizes physiological data to predict mortality, 30-days readmission, and ICU admissions.  & \begin{itemize} \item Not personalized. Uses vital signs and lab tests to construct a ``one-size-fits" all population-level model. \item Ignores correlations between vital signs, and hence may double-count risk factors (see Eq. (1) in \cite{rothman2013development}). \item Uses the instantaneous vital signs and lab tests measurements, and ignores the physiological stream trajectory. \end{itemize} \\
		&  &  & \\
    \hline
		\hline
  \end{tabular}
	\captionsetup{font= small}
  \caption{Summary of the state-of-the-art critical care risk scores.}
\end{table*}
   
\subsection{Summary of Contributions} 

We develop a risk scoring algorithm that provides real-time, personalized assessments for the acuity of critical care patients in a hospital ward. The algorithm is trained using the EHR data in an offline stage, and risk scores for a newly hospitalized patient are computed via the trained model in real-time using her temporal, irregularly sampled physiological data, which resemble the data structure depicted in Fig. \ref{prevwfig}. The proposed risk score has the following features:  
\begin{itemize}
\item The patient's physiological streams are modeled using a generative multitask Gaussian Process (GP) \cite{durichen2015multitask},\cite{ghassemi2015multivariate}, the parameters of which depend on the patient's clinical status, i.e. whether the patient is clinically stable or deteriorating. We capture the non-stationarity of the deteriorating patients' physiological streams by dividing every patient's stay in the ward into a sequence of temporal {\it epochs}, and allow the parameters of the multitask GP to vary across these epochs. Non-stationarity is taken into account in the training phase by temporally aligning the physiological streams recorded in the EHR data, and is taken into account in the real-time deployment phase by continuously estimating the multitask GP epoch index over time. \\
   
\item The patient's risk score is computed as the optimal test statistic of a sequential hypothesis test that disentangles clinically stable patients from the clinically deteriorating ones as more physiological measurements are gathered over time. Our conception of the risk score follows the seminal work of Wald on sequential analysis \cite{wald1973sequential}.\\

\item The heterogeneity of the patients' population is captured by considering the patients' latent subtypes (or {\it phenotypes} \cite{saria2015subtyping}). The proposed algorithm discovers the number of patient subtypes from the training data, and learns a separate multitask GP model for the physiological streams associated with each subtype. Discovering the patients' latent subtypes is carried out using unsupervised learning (the expectation-maximization (EM) algorithm) over the domain of clinically stable patients since these patients are dominant in the dataset (i.e. they comprise more than $90\%$ of the EHR records), and are more likely to exhibit stationary physiological trajectories, thus their physiological streams are described with few hyper-parameters and can be efficiently estimated. \\  

\item The knowledge of the patients' latent subtypes which was extracted from the domain of clinically stable patients is then transferred to the domain of clinically deteriorating patients via {\it self-taught} transfer learning, where the algorithm learns a set of GP models for the different classes of clinically deteriorating patients. Every GP model associated with (stable or deteriorating) patients who belong to a specific subtype is called a {\it GP expert}. Thus, every GP expert specialized in scoring the risk for one of the discovered patient subtypes. \\ 

\item For a newly hospitalized patient, the posterior beliefs of all GP experts about the patient's clinical status given her physiological data stream are computed and updated in real-time, and the risk score is evaluated as a weighted average of those belief processes. The weights are computed based on the patient's hospital admission information, and are derived from the probability of the patient's membership in each of the discovered subtypes as a function of her admission information (e.g. age, gender, transfer status, transplant status, etc), which we estimate using {\it transductive transfer learning}. \\
\end{itemize}

Experiments were conducted using a dataset for a heterogeneous cohort of 6,321 patients who were admitted during the years 2013-2016 to a general medicine floor in the Ronald Reagan UCLA medical center, a tertiary medical center. The proposed risk scoring model was trained using 5,130 patients, and tested for the most recently admitted 1,191 patients in the cohort (admitted during the years 2015-2016). Results show that the proposed risk score consistently outperforms the Rothman index, MEWS, APACHE and SOFA scores, in terms of timeliness and accuracy (i.e. the true positive rate (TPR) and the positive predictive value (PPV)), in addition to state-of-the-art machine learning algorithms such as random forest, LASSO, logistic regression, etc. The results show that the proposed risk score boosts the AUC with 12$\%$ as compared to the Rothman index ($p$-value $<$ 0.01), and can prompt alarms for ICU admission 12 hours before clinicians (on average) for a PPV of 25$\%$ and TPR of 50$\%$, which provides the ward staff with a safety net for patient care by giving them sufficient time to intervene at an earlier time in order to prevent clinical deterioration. Moreover, the proposed risk score reduces the number of false alarms per number of true alarms for any setting of the TPR, which reduces the alarm fatigue and allows for better hospital resource management. We also provide some clinical insights by highlighting the number of discovered patient subtypes, and the admission information that are relevant to subtype discovery. 

\subsection{Related Works}
\label{sub1.3}
Two broad categories of risk models and scores that quantify a patient's risk for an adverse event have been developed in the medical literature. The first category comprises {\it early-warning scores} (EWS), which hinge on expert-based models for triggering transfer to ICU \cite{morgan1997early}. Notable examples of such scores are MEWS and its variant VitalPAC \cite{prytherch2010views}. These scores rely mainly on experts to specify the risk factors and the risk scores associated with these factors \cite{subbe2001validation}. A major drawback of this class of scores is that since the model construction is largely relying on experts, the implied risk functions that map physiological parameters to risk scores do not have any rigorous validation. Recent systematic reviews have shown that EWS-based alarm systems only marginally improve patient outcomes while substantially increasing clinician and nursing workloads \cite{tsien1997poor, cvach2012monitor, bliss2000behavioural}. Other expert-based prognostication scores that were constructed to predict mortality in the ICU, such as SOFA and APACHE scores, has been shown to provide a reasonable predictive power when applied to predict deterioration for patients in wards \cite{yu2014comparison}. \\ 

The second category of risk scores relies on more rigorous, data-intensive regression models to derive and validate risk scoring functions using the electronic medical record. Examples for such risk scores include the regression-based risk models developed by Kirkland et al. \cite{kirkland2013clinical}, and by Escobar et al. \cite{escobar2012early}. Rothman et al. build a more comprehensive model for computing risk scores on a continuous basis in order to detect a declining trend in time \cite{rothman2013development, finlay2014measuring}. The risk score computed therein, which is termed as the ``Rothman index", quantifies the individual patient condition using 26 clinical variables (vital signs, lab results, cardiac rhythms and nursing assessments). Table I summarizes the state-of-the-art risk scores used for critical care prognostication.\\  
   
The Rothman index is the state-of-the-art risk scoring technology for patients in wards: about 70 hospitals and health-care facilities, including Houston Methodist hospital in Texas, and Yale-New Haven hospital in Connecticut, are currently deploying this technology \cite{WSJ}. While validation of the Rothman index have shown its superiority to MEWS-based models in terms of false alarm rates \cite{finlay2014measuring}, the risk scoring scheme used for computing the Rothman index adopts various simplifying assumptions. For instance, the risk score computed for the patient at every point of time relies on instantaneous measurements, and ignores the history of previous vital sign measurements (see Equation (1) in \cite{rothman2013development}). Moreover, correlations among vital signs are ignored, which leads to double counting of risk factors. Finally, the Rothman scoring model is fitted to provide a reasonable ``average" predictive power for the whole population of patients, but does not offer ``personalized" risk assessments for individual patients, i.e. it ignores baseline and demographic information available about the patient at admission time. Our risk scoring model addresses all these limitations, and hence provides a significant gain in the predictive power as compared to the Rothman index as we show in Section IV. \\

The problem of modeling temporal physiological data was previously considered by the machine learning and data mining communities. Physiological models that rely on multitask GPs were previously considered in \cite{clifton2012gaussian, ghassemi2015multivariate, durichen2015multitask, pimentel2013modelling, schulam2015framework}. In these works, the focus was to predict the futuristic vital signs and lab tests values via GP regression (e.g. estimating future values of the Cerebrovascular pressure reactivity in \cite{ghassemi2015multivariate}), and the quality of predictions was assessed using metrics such as the mean-square error. Our work departs from this strand of literature in many ways. First, our goal is to infer the patient's latent status given the evidential vital signs and lab tests data using the multitask GP, and hence we need to deal with different types of patients with different physiological models rather than train a single disease progression model for a population of patients that has a specific chronic disease \cite{wang2014unsupervised}. Second, clinically deteriorating patients do exhibit a non-stationary physiological behavior, and hence the models in \cite{clifton2012gaussian, ghassemi2015multivariate, durichen2015multitask, pimentel2013modelling}, which have been reliant on the stationary {\it squared-exponential} covariance kernel to construct the GPs, would not suffice as a reliable model for the patients' physiological streams. Finally, previous physiological models (with the exception of that in \cite{schulam2015framework}) were constructed in a ``one-size-fits-all" fashion, i.e. the hyper-parameters are tuned independent of personal and demographic features of the individual patients, and the same model is shared among the entire patients' population. \\  

Various other important tools for risk prognosis that do not rely on GP models have been recently developed. In \cite{henry2015targeted} and \cite{dyagilev2015learning}, a Cox regression-based model was used to develop a sepsis shock severity score that can handle data streams that are censored due to interventions. However, this approach does not account for personalization in its severity assessments, and relies heavily on the existence of ordered pairs of comparisons for the extent of disease severity at different times, which may not always be available and cannot be practically obtained from experts. Our model does not suffer from such limitations: it does not rely on proportional hazard estimates, and hence does not require ordered pairs of disease severity temporal comparisons, and can be trained using the raw physiological stream records that are normally fed into the EHR during the patients' stay in the ward. Personalized risk prognosis models were developed in \cite{ng2015personalized, schulam2015framework, wang2015towards} and \cite{visweswaran2010learning}. \\

In \cite{ng2015personalized} and \cite{wang2015towards}, personalized risk factors are computed for a new patient by constructing a dataset of $K$ ``similar patients" in the training data, and train a predictive model for that patient. This approach would be computationally very expensive when applied in real-time for patients in a ward since it requires re-training a model for every new patient, and more importantly, it does not recognize the extent of heterogeneity of the patients, i.e. the constructed dataset has a fixed size of $K$ irrespective of the underlying patients' physiological heterogeneity. Hence, such methods may incur efficiency loss if $K$ is underestimated, and may perform unnecessary computations if the underlying population is already homogeneous. Our model overcomes this problem by learning the number of latent subtypes from the data, and hence it can adapt to both homogeneous and heterogeneous patient populations. \\

The rest of the paper is organized as follows. In Section II, we present a physiological model for the vital signs and lab tests of hospitalized patients in wards. In Section III, we propose a risk scoring algorithm that efficiently learns the parameters of the model presented in Section II, and computes risk scores for hospitalized patients in real-time. Experiments on a real-world dataset are conducted in Section IV, and the paper is concluded in Section V.

\section{The Physiological Model}
In this section, we present a comprehensive model for the patients' physiological data and develop a rigorous formulation for the risk scoring problem. A risk scoring algorithm that utilizes the model presented hereunder is developed in the next Section.

\begin{figure*}[t!]
    \centering
    \includegraphics[width=6.5 in]{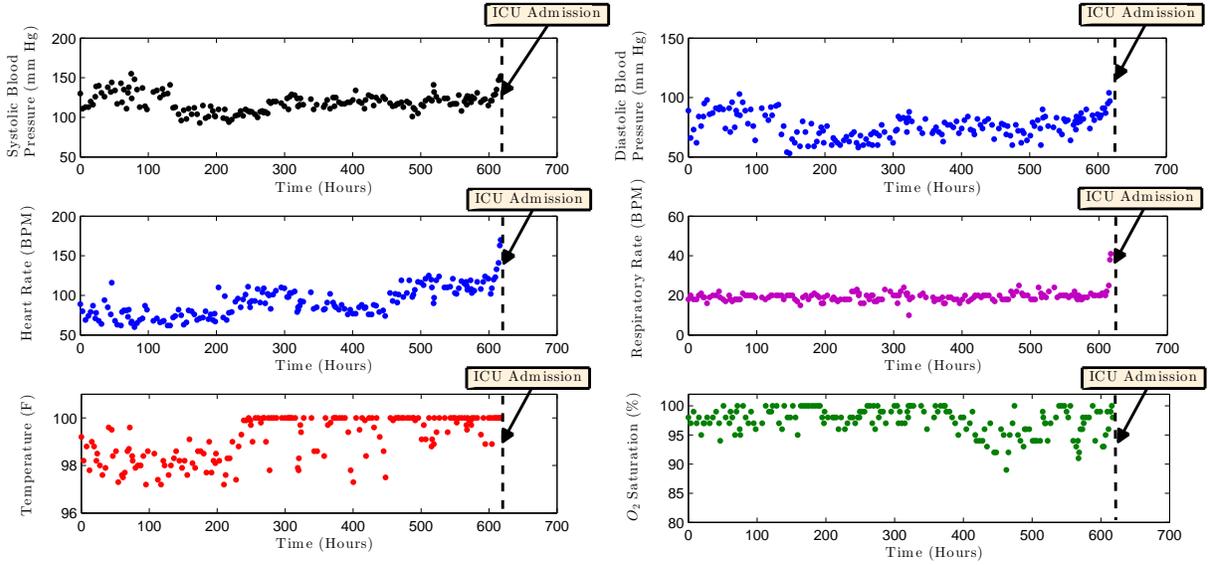}  
		\captionsetup{font= small}	
    \caption{Physiological streams of a patient hospitalized in a ward for 620 hours and then admitted to ICU upon her clinical deterioration.}
		\label{Fiq1}
\end{figure*}

\subsection{Modeling the Patients' Risks and Clinical Status}
Two types of information are associated with every patient in the (surgical or medical) ward:\\
\\
{\bf 1- Physiological information $X(t)$:} We define $X(t) = [X_{1}(t), X_{2}(t), .\,.\,., X_{D}(t)]^{T}$ as a $D$-dimensional stochastic process representing the patient's $D$ physiological streams (lab tests and vital signs) as a function of time. The process $X_i(t)$ takes values from a space $\mathcal{X}_i$, and $\mathcal{X} = \mathcal{X}_{1} \times \mathcal{X}_{2} \times .\,.\,., \times \mathcal{X}_{D}$. Vital signs and lab tests are gathered at arbitrary time instances $\{t_{ij}\}_{i=1,j=1}^{D, M_{i}}$ (where $t=0$ is the time at which the patient is admitted to the ward), where $M_i$ is the total number of samples of vital sign (or lab test) $i$ that where gathered during the patient's stay in the ward. Thus, the set of all observations of the physiological data that the ward staff has for a specific patient is given by $\{X_i(t_{ij})\}_{i=1,j=1}^{D, M_{i}}$, and we will refer to the realizations of these variables as $\{x_{ij}, t_{ij}\}_{ij}$.\\
\\
{\bf 2- Admission information $Y$:} We define the $S$-dimensional random vector $Y$ as the patient's static information obtained at admission (e.g. age, gender, ICD9 code, etc). The random vector $Y$ is drawn from a space $\mathcal{Y}$, and we denote the realizations of the patient's static information as $Y = y$. Thus, the set of all (static and time-varying) information associated with a patient can be gathered in a set $\{y,\{x_{ij}, t_{ij}\}_{ij}\}$. \\ 
\\
Fig. \ref{Fiq1} depicts the vital signs (Systolic blood pressure, diastolic blood pressure, heart rate, respiratory rate, temperature and $O_2$ saturation) gathered during (at irregularly spaced time instances) during the stay of a critically ill patient in a ward (around 620 hours), before being admitted to the ICU by the ward staff members who have observed her deteriorating clinical status. In this example, the set $\{x_{ij}, t_{ij}\}_{ij}$ contains the vital sign measurements and their respective sampling signs.\\  

Let $V \in \{0,1\}$ be a binary latent variable that corresponds to the patient's true clinical status; 0 standing for a stable clinical status, and 1 for a clinically deteriorating status. Since physiological streams manifest the patients' clinical statuses, it is natural to assume that the conditional distributions of $X^{o}(t) = X(t)\left|V=0\right.$ differ from that of $X^{1}(t) = X(t)\left|V=1\right.$. We assume that $V$ is drawn randomly for every patient at admission time and stays fixed over the patient's stay in the ward, i.e. the value of $V$ is revealed at the end of every physiological stream, where $V=1$ if the patient is admitted to the ICU, and $V=0$ if the patient is discharged home. During the patient's stay in the ward, the ward staff members are confronted with two hypotheses: the null hypothesis $\mathcal{H}_{o}$ corresponds to the hypothesis that the patient is clinically stable, whereas the alternative hypothesis $\mathcal{H}_{1}$ corresponds to the hypothesis that the patient is clinically deteriorating, i.e.   
\begin{equation}
V = 
\left\{
\begin{array}{ll}
      0: \,\, \mathcal{H}_{o} \,\, (\mbox{clinically stable patient}),\\
      1: \,\, \mathcal{H}_{1} \,\, (\mbox{clinically deteriorating patient}).\\
\end{array} 
\right.
\label{eqq1} 
\end{equation}
Thus, the prognosis problem is a {\it sequential hypothesis test} \cite{wald1973sequential}, i.e. the clinicians need to reject one of the hypotheses at some point of time after observing a series of physiological measurements. Hence, following the seminal work of Wald in \cite{wald1973sequential}, we view the patient's risk score as the test statistic of the sequential hypothesis test. That is, the patient's risk score at time $t$, which we denote as $\bar{R}(t)\in [0,1]$, is the posterior probability of hypothesis $\mathcal{H}_{1}$ given the observations $\{x_{ij}, t_{ij}\leq t\}_{ij}$, and we have that  
\begin{align}
\bar{R}(t) &= \mathbb{P}\left(\mathcal{H}_{1}\left|\{x_{ij}, t_{ij}\leq t\}_{ij}\right.\right) \nonumber \\
&= \frac{\mathbb{P}\left(\{x_{ij}, t_{ij}\leq t\}_{ij}\left|\mathcal{H}_{1}\right.\right) \cdot \mathbb{P}\left(\mathcal{H}_{1}\right)}{\sum_{v \in \{0,1\}}\mathbb{P}\left(\{x_{ij}, t_{ij}\leq t\}_{ij}\left|\mathcal{H}_{v}\right.\right) \cdot \mathbb{P}\left(\mathcal{H}_{v}\right)},
\label{eqq2} 
\end{align}
where $\mathbb{P}\left(\mathcal{H}_{1}\right)$ is the prior probability of a patient in the ward being admitted to the ICU (i.e. the rate of ICU admissions). 

\subsection{Modeling the Physiological Signals}
Since the vital signs and lab tests are gathered at arbitrary, irregularly sampled time instances, it is convenient to adopt a continuous-time model for the patients' physiological stream using GPs \cite{durichen2015multitask, bonilla2007multi, ghassemi2015multivariate}. We model the $D$ (potentially correlated) physiological streams of a monitored patient as a multitask GP defined over $t \in \mathbb{R}_{+}$. The model parameters depend on the patient's latent clinical status $V$. Since clinically stable patients do not exhibit changes in their clinical status, we adopt a stationary model for $X^{o}(t)$. Contrarily, deteriorating patients pass through phases of clinical acuity, which invokes the need for a non-stationary model for $X^{1}(t)$. In the following, we present the physiological models for clinically stable and deteriorating patients, which we will then use as a proxy for risk scoring in the next Section. \\
\\
{\bf Physiological Signals Model for Clinically Stable Patients}\\ 
\\
For clinically stable patients, i.e. $V=0$, we adopt a multitask GP model for the physiological signal $X^{o}(t)$ as follows 
\begin{equation}
X^{o}(t) \sim \mathcal{GP}(m_o(t),k_o(i,j,t,t^{'})),
\label{eqq3} 
\end{equation}
where $m_o(t): \mathbb{R}^{+} \rightarrow \mathcal{X}$ is the {\it mean function}, and $k_o(i,j,t,t^{'}): \mathcal{X}_i \times \mathcal{X}_j \times \mathbb{R}^{+} \times \mathbb{R}^{+} \rightarrow \mathbb{R}_{+}$ is the {\it covariance kernel}. The mean function is assumed to be a constant vector, i.e. $m_o(t) = [m^{1}_o, m^{2}_o,.\,.\,.,m^{D}_o]^{T},$ the entries of which represent the average value of the different physiological streams (e.g. the mean value of the respiratory rate depicted in Fig. \ref{Fiq1} is 20). We assume that the covariance kernel matrix $k_o(i,j,t,t^{'})$ has the following separable form  
\begin{align}
k_o(i,j,t,t^{'}) = {\bf \Sigma}_o(i,j)\,k_o(t,t^{'}),
\label{eqq4}
\end{align}
where ${\bf \Sigma}_o$ is a {\it stationary correlation matrix} that quantifies the correlations between the various physiological streams. The kernel function $k_o(t,t^{'})$ is {\it squared-exponential} kernel \cite{rasmussen2006gaussian, clifton2013gaussian, durichen2015multitask}, defined as 
\begin{equation}
k_o(t,t^{'}) = \omega_{o}^{2} \, e^{-\frac{1}{2\ell_o^{2}}\,||t-t^{'}||^{2}},
\label{eqq5}
\end{equation} 
where $\omega_{o}$ and $\ell_o$ are hyper-parameters: $\omega_{o}$ is the {\it variance hyper-parameter}, and $\ell_o$ is the {\it characteristic length-scale}. The parameter $\omega_{o}$ controls the dynamic range of the fluctuations of $X(t)$; the parameter $\ell_o$ controls the rate of such fluctuations. Note that (\ref{eqq4}) implies that we assume that all the physiological streams have the same temporal characteristics, i.e. the same variance and characteristic length-scale. \\

Since the correlation matrix ${\bf \Sigma}_o$ needs to be positive semi-definite, we adopt the ``free-form" construction of the correlation matrix via the Cholesky decomposition as follows 
\begin{equation}
{\bf \Sigma}_o = {\bf L}_o\,{\bf L}_o^{T},\, {\bf L}_o = \begin{bmatrix}
    \sigma_{o,1} & 0 & \dots  & 0 \\
		\sigma_{o,2} & \sigma_{o,3} & \dots  & 0 \\
    \vdots & \vdots & \ddots & \vdots \\
    \sigma_{o,\bar{D}-m+1} & \sigma_{o,\bar{D}-m+2} & \dots  & \sigma_{o,\bar{D}}
\end{bmatrix},
\label{eqq6}
\end{equation}
where $\bar{D} = \frac{D(D+1)}{2}$ \cite{bonilla2007multi}. Since the variance of each stream is already captured by the entries of ${\bf \Sigma}_o$, we assume that $\omega_{o} = 1$ for all streams. Thus, the hyper-parameters that characterize a multi-task GP $\mathcal{GP}(m_o(t),k_o(i,j,t,t^{'}))$ are $\ell_o$ and the entries of ${\bf L}_o$, which we compactly write in a vector ${\bf \sigma}_o$ as follows 
\begin{equation}
{\bf \sigma}_o = \left[\sigma_{o,1}, .\,.\,., \sigma_{o,\bar{D}-1}, \sigma_{o,\bar{D}}\right].
\label{eqq7}
\end{equation}
We summarize the parameters of the GP model capturing the physiological streams of clinically stable patients via the following parameter set
\begin{equation}
{\bf \Theta}_{o} = \{\{m^{d}_{o}\}_{d=1}^{D}, \,\, \ell_{o}, \,\, {\bf \sigma}_{o}\},
\label{eqq8}
\end{equation}
which aggregates the $\frac{D(D+1)}{2}+D+1$ hyper-parameters of the multi-task GP. We write $X^{o}(t) \sim \mathcal{GP}({\bf \Theta}_{o})$ to denote an instance of a physiological stream of a clinically stable patient generated with a parameter set ${\bf \Theta}_{o}$. \\
\\
{\bf Physiological Signals Model for Clinically Deteriorating Patients}\\ 
\\
For clinically deteriorating patients, i.e. patients with $V = 1$, we adopt a non-stationary model for $X^{1}(t)$ specified as follows
\begin{equation}
X^{1}(t) \sim \mathcal{GP}({\bf \Theta}_{1}),
\label{eqq9} 
\end{equation}
where ${\bf \Theta}_{1}$ is the parameter set for the physiological streams of deteriorating patients. Since deteriorating patients exhibit changes in their clinical status (e.g. progression from a more stable status to a less stable one), a stationary covariance kernel, such as the one defined in (\ref{eqq5}), and a constant mean function do not suffice to describe the physiological stream of a deteriorating patient. For instance, we can see that the temperature measurements' stream in Fig. \ref{Fiq1} exhibit a change in its mean and variance characteristics after a stay of 250 hours in the ward. This motivates a non-stationary model for $X^{1}(t)$ that divides the time domain into a sequence of {\it epochs}, each is of duration $T_1$, and is associated with a distinct constant mean function and a distinct squared-exponential covariance kernel. \\    

Let $T = K \cdot T_1$ be the maximum duration for a patient's stay in the ward. That is, the patient passes through $K$ consecutive epochs, each of which has a mean function and a covariance kernel parametrized by ${\bf \Theta}^{k}_{1} = \{\{m^{d}_{1,k}\}_{d=1}^{D}, \,\, \ell_{1,k}, \,\, {\bf \sigma}_{1,k}\}, \forall k \in \{1,2,.\,.\,.,K\}$. Since patients arrive at the hospital ward at random time instances, at which the clinical status is unknown, we define $\bar{k} \in \{1,2,.\,.\,.,K\}$ as the unobservable, initial epoch index, which we assume to be drawn from an unknown distribution $\bar{k} \sim f_k(k)$. The physiological measurements gathered by the clinicians during the patient's are governed by a monotonically increasing sequence of epochs, i.e. the clinicians observe physiological measurements drawn from a process with the underlying epoch sequence $\{\bar{k}, \bar{k}+1,.\,.\,.,K\}$. For instance, if $K=6$ and the realization of $\bar{k}$ is 3, then the (deteriorating) patient's physiological process $X^{1}(t)$ has its parameters changing over time according to the epoch sequence $\{3,4,5,6\}$. Note that the length of the patient's stay in the ward is given by $(K-\bar{k}+1) \cdot T_1$, which is random since $\bar{k}$ is a random variable. \\  

\begin{figure}[t!]
    \centering
    \includegraphics[width=3.5 in]{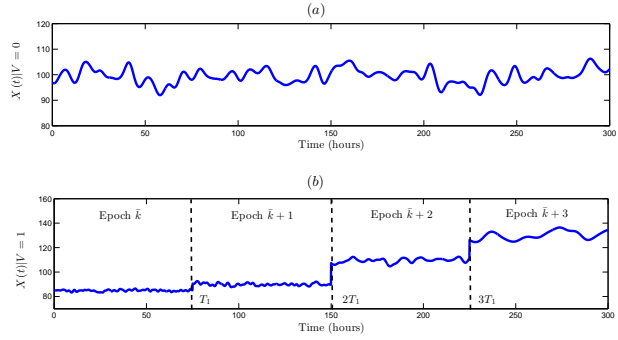}    
		\captionsetup{font= small}
    \caption{Exemplary sample paths for $X^{o}(t)$ (Fig. 2(a)) and $X^{1}(t)$ (Fig. 2(b)).}
		\label{Fiq2}
\end{figure}

We assume that the physiological measurements across different epochs are independent, but measurements within the same epoch are correlated. Thus, the vital signs and lab tests are correlated within every interval in the set of intervals $\{[0,T_1), [T_1,2T_1),.\,.\,., [(K-\bar{k})\,T_1, (K-\bar{k}+1)\,T_1)\}$, but are uncorrelated across different time intervals. In other words, the covariance kernel for the process $X^{1}(t)$ is given by   
\begin{align}
k_1(i,j,t,t^{'}) = \left\{
\begin{array}{ll}
     {\bf \Sigma}_{1,k}(i,j)\,k_{1,k}(t,t^{'}), \,\, \forall t,t^{'} \in [t_1, t_2),\\
      0, \,\, \mbox{Otherwise}
\end{array} 
\right.
\label{eqq10}
\end{align}
where $[t_1, t_2) \in \{[0,T_1),.\,.\,., [(K-\bar{k})\,T_1, (K-\bar{k}+1)\,T_1)\}$, and
\begin{equation}
k_{1,k}(t,t^{'}) = \omega_{1,k}^{2} \, e^{-\frac{1}{2\ell^{2}_{1,k}}\,||t-t^{'}||^{2}}.
\label{eqq11}
\end{equation} 
The parameters of the GP model for deteriorating patients can be summarized via the following parameter set
\begin{equation}
{\bf \Theta}_{1} = \{\{m^{d}_{1,k}\}_{d=1}^{D}, \,\, \ell_{1,k}, \,\, {\bf \sigma}_{1,k}\}_{k=1}^{K}.
\label{eqq12}
\end{equation}  
The parameter set ${\bf \Theta}_{1}$ encapsulates $K\left(\frac{D(D+1)}{2}+D+1\right)$ hyper-parameters that describe the process $X^{1}(t)$. Note that the model $X^{1}(t)$ entails much more parameters than the model $X^{o}(t)$, which poses a significant challenge in learning the parameters of $X^{1}(t)$. We address this challenge elaborately in the next Section. Fig. \ref{Fiq2} illustrates a sample path from the process $X^{o}(t)$ and a sample path from $X^{1}(t)$, highlighting the differences between the two generative models.      

\subsection{Modeling Patients' Subtypes}

\begin{figure*}[t!]
    \centering
    \includegraphics[width=5.5 in]{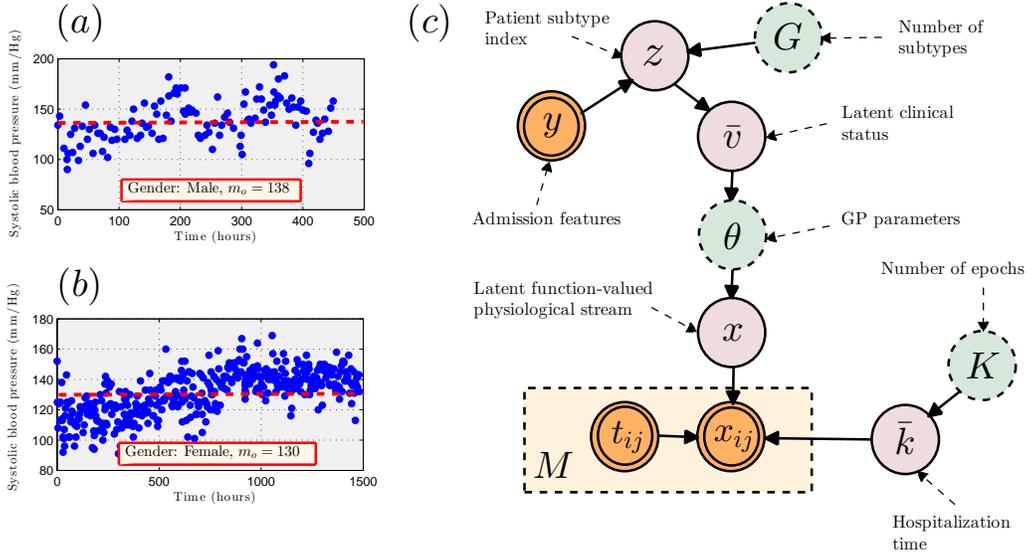} 
		\captionsetup{font= small}		
    \caption{A depiction for: (a) the systolic blood pressure signal for a clinically stable male. The mean function for clinically stable males is $m_o = 138$ mm/Hg, (b) the systolic blood pressure signal for a clinically stable female. The mean function for clinically stable females is $m_o = 130$ mm/Hg, and (c) a graphical model for the patients' physiological signals' generative process (observed variables are double-edged, and model parameters are presented with dotted edges.).}
		\label{Fiq3}
\end{figure*}

The model presented so far is constructed in a ``one-size-fits-all" fashion. That is, the risk score computed in (\ref{eqq2}) considers the vital signs and lab tests for the monitored patient, without considering her baseline admission information (the vector $Y$). The interpretation of the manifest variables $\{x_{ij}, t_{ij}\}_{ij}$ in terms of the risk for clinical deterioration may differ depending on the patient's age, gender, transfer status, or clinical history. Thus, a risk score that is tailored to the individual's admission feature would ensure a higher level of granularity in modeling the physiological signals, which would lead to a more accurate prognosis. \\

In order to ensure that our risk score is ``personalized", we model the heterogeneity of the patients' population by incorporating a {\it subtype} variable $Z \in \mathcal{Z} = \{1,2,.\,.\,.,G\}$, which indicates the patient's latent {\it phenotype} which determines her physiological behavior, where $G$ is the number of subtypes to which a patient may belong. That is, every patient has her physiological behavior being determined by both her clinical status and her latent subtype. We denote risk scores that take the patient's particular subtype into account as {\it ``personalized risk scores}".  \\

The influence of the patient's subtype $Z$ on the patient's physiological model is captured by the following relations
\begin{align}
Z \perp V &| Y, \nonumber \\
V \perp Y &| Z,
\label{eqq13}
\end{align}
where $\perp$ denotes conditional independence. The relations in (\ref{eqq13}) imply that: (a) a patient's subtype is independent of her clinical status given her admission information, and (b) a patient's clinical status is independent of the admission information given her subtype. That is, knowledge of the patent's admission information suffices to infer her subtype (e.g. knowledge of age and gender, etc, is enough to know the subtype to which a patient belongs irrespective of the true clinical status), and knowledge of the patient's subtype is enough to infer the patient's vulnerability irrespective to the admission information. The first relation follows from the fact that the patient's subtype is an intrinsic feature of the patient that is independent of her clinical acuity, whereas the second relation follows from that fact that the information contained in $Y$ is a subset of the information contained in the patient's intrinsic subtype $Z$.  \\

The patient's subtype manifests in her physiological signals by manipulating the parameter sets for the multitask GPs representing both $X^{o}(t)$ and $X^{1}(t)$. In other words, the parameters of the multitask GP modeling the patient's physiological signal depends not only on her clinical status $V$, but also on her subtype $Z$. The parameter set for clinically deteriorating patients is denoted as ${\bf \Theta}^{z}_{1}$, and the parameter set for stable patients is denoted as ${\bf \Theta}^{z}_{o}$, where $Z=z$ is a realization for the patient's subtype. The construction of both parameter sets follows the description provided in the previous subsection. Therefore, the physiological signals for the patients in the ward are generated as follows
\begin{equation}
X^{v}(t)\left|Z=z\right. \sim \mathcal{GP}({\bf \Theta}^{z}_{v}).
\label{eqq14} 
\end{equation}

Fig. \ref{Fiq3}(a) and \ref{Fiq3}(b) depict a particular physiological stream (systolic blood pressure) for a clinical stable male (Fig. \ref{Fiq3}(a)) and a clinically stable female (Fig. \ref{Fiq3}(b)). It can be seen that even though both patients share the same clinical status, this status manifests differently for the two patients, i.e. the average systolic blood pressure for males is higher than that for females. We can also see that the variance of the measurements is higher for the male's signal. This indicates the necessity of incorporating the information in $Y$ while assessing the patient's risk, since otherwise the risk maybe overestimated or underestimated for the patient leading to either a delayed or an unnecessary ICU transfer. \\

Fig. \ref{Fiq3}(c) depicts a graphical model describing the generative process for the patients' physiological signals. The patient's subtype $Z=z$ is hidden, and affects both her clinical status $V = v$ and the physiological behavior that manifests in the vital signs and lab tests. The variable $\bar{V} \in \{0,1\} \times \mathcal{Z}, \bar{V} = [V \,\, Z]^{T}$ augments both the patient's subtype and clinical status; a realization of this variable $\bar{V} = \bar{v}$ determines the parameter set $\theta | \bar{v} = {\bf \Theta}^{z}_{v},$ which is used to generate a latent function-valued variable $X(t) = x \in \mathbb{R}^{\mathbb{R}^{D}}$. A plate model is then used to describe the sequence of measurements $\{x_{ij}\}_{i,j}$ gathered by the clinicians at time instances $\{t_{ij}\}_{i,j}$. The time instances $\{t_{ij}\}_{i,j}$ are assumed to be exogenously determined by the ward staff and are uninformative of the clinical status, hence they are modeled as parent nodes in the graphical models. Observations are influenced by the index of the first epoch, $\bar{k}$, which is also assumed to be exogenously determined by the patient's arrival to the ward. It can be seen that the probabilistic influences among the variables $V$, $Z$ and $Y$ in the graphical model in Fig. \ref{Fiq3}(c) capture the relations specified in (\ref{eqq13}). \\

Having defined the patients' subtypes, we refine the definition of the (non-personalized) risk score $\bar{R}(t)$, and incorporate the patient's individual static features in a personalized risk score $R(t,y)$ as follows
\begin{align}
R(t,y) &= \mathbb{P}\left(\mathcal{H}_{1}\left|\{x_{ij}, t_{ij}\}_{i,j}, Y = y\right.\right) \nonumber \\
&= \sum_{z \in \mathcal{Z}}\mathbb{P}\left(\mathcal{H}_{1}\left|\{x_{ij}, t_{ij}\}_{i,j}, Z = z\right.\right) \cdot \mathbb{P}(Z=z | Y = y) \nonumber \\
&= \sum_{z \in \mathcal{Z}}\mathbb{P}\left(V=1\left|\{x_{ij}, t_{ij}\}_{i,j}, {\bf \Theta}^{z}_{o}, {\bf \Theta}^{z}_{1}\right.\right) \cdot \mathbb{P}(Z=z | Y = y),
\label{eqq15} 
\end{align}   
where
\[\mathbb{P}\left(V=1\left|\{x_{ij}, t_{ij}\}_{i,j}, {\bf \Theta}^{z}_{o}, {\bf \Theta}^{z}_{1}\right.\right) = \]
\begin{align}
\frac{\mathbb{P}\left(\{x_{ij}, t_{ij}\}_{i,j}\left|{\bf \Theta}^{z}_{1}\right.\right) \cdot \mathbb{P}(V=1|Z=z)}{\sum_{v \in \{0,1\}}\mathbb{P}\left(\{x_{ij}, t_{ij}\}_{i,j}\left|{\bf \Theta}^{z}_{v}\right.\right) \cdot \mathbb{P}(V=v|Z=z)},
\label{eqq16} 
\end{align} 
where we have assumed in (\ref{eqq15}) and (\ref{eqq16}) that the epoch index $\bar{k}$ is observed and we dropped the conditioning on $\bar{k}$ for simplicity of exposition. In the next Section, we develop an algorithm that learns the patients' physiological model from offline data, and computes the monitored patients' personalized risk scores using (\ref{eqq15}) and (\ref{eqq16}).

\section{A Personalized Risk Scoring Algorithm}
In this Section, we propose an algorithm that learns the physiological model presented in the previous Section from offline data, and computes the risk score formulated in (\ref{eqq15}) and (\ref{eqq16}) for newly hospitalized patients in real-time.

\subsection{Objectives}
Given an offline training dataset $\mathcal{D}$ that comprises $N$ {\it reference patients} whose physiological measurements were recorded in the electronic health record (EHR), we aim at learning a personalized risk scoring model, i.e. learning the parameters of the model presented in Section II, and applying the learned risk model for newly hospitalized patients.\\  

The training dataset $\mathcal{D}$ is represented as a collection of tuples
\[\mathcal{D} = \left\{\left(\{x^{(n)}_{ij}, t^{(n)}_{ij}\}_{i,j}, y^{(n)}, v^{(n)}\right)\right\}_{n=1}^{N},\] 
where each element in $\mathcal{D}$ corresponds to a reference patient; $\{x^{(n)}_{ij}, t^{(n)}_{ij}\}_{i,j}$ is the set of vital signs and lab tests measurements, $y^{(n)}$ is the admission information, and $v^{(n)}$ is the true clinical status (i.e. patient is admitted to the ICU or discharged home) of the $n^{th}$ patient in $\mathcal{D}$. For $v \in \{0,1\},$ let 
\[\mathcal{D}_{v} = \left\{\left(\{x^{(n)}_{ij}, t^{(n)}_{ij}\}_{i,j}, y^{(n)}, v^{(n)}\right): v^{(n)} = v\right\},\]
where $\mathcal{D}_{o}$ is the set of data points for clinically stable patients, and $\mathcal{D}_{1}$ is the set of data points for clinically deteriorating patients, and $N_v = |\mathcal{D}_{v}|$ is the size of the dataset $\mathcal{D}_{v}$. \\
\\

Our algorithm $\mathcal{A}$ operates in two modes: an {\it offline mode} $\mathcal{A}_{off}$, in which a risk scoring model is learned from the offline dataset $\mathcal{D}$, and an {\it online mode} $\mathcal{A}_{on}$, in which a risk score is sequentially computed for a newly hospitalized patient with a sequence of physiological measurements $\{x_{ij}, t_{ij}\}_{i,j}$, i.e.
\[(\hat{{\bf \Theta}}^{1}_{o},.\,.\,.,\hat{{\bf \Theta}}^{G}_{o},\hat{{\bf \Theta}}^{1}_{1},.\,.\,.,\hat{{\bf \Theta}}^{G}_{1}) = \mathcal{A}_{off}(\mathcal{D}),\]  
\[R(t,y) = \mathcal{A}_{on}(\{x_{ij}, t_{ij}\leq t\}_{i,j}, \hat{{\bf \Theta}}^{1}_{o},.\,.\,.,\hat{{\bf \Theta}}^{G}_{o},\hat{{\bf \Theta}}^{1}_{1},.\,.\,.,\hat{{\bf \Theta}}^{G}_{1}).\]
That is, $\mathcal{A}_{off}$ estimates the parameter set for stable and deteriorating patients for all subtypes $(\hat{{\bf \Theta}}^{1}_{o},.\,.\,.,\hat{{\bf \Theta}}^{G}_{o},\hat{{\bf \Theta}}^{1}_{1},.\,.\,.,\hat{{\bf \Theta}}^{G}_{1})$, whereas $\mathcal{A}_{on}$ implements (\ref{eqq15}) and (\ref{eqq16}) to assign a risk score for the monitored patient in real-time. \\  

In order to evaluate the predictive power of the algorithm $\mathcal{A}$, we set a threshold $\eta$ on the computed risk score $R(t,y)$, and allow the algorithm to prompt an alarm (i.e. declare the hypothesis $\mathcal{H}_{1}$) whenever the risk score crosses that threshold. This resembles the structure of the optimal sequential hypothesis test, where the null hypothesis is rejected whenever the test statistic crosses a predefined threshold \cite{wald1973sequential}. We define $T_s$ as the {\it stopping time} at which the risk score computed by the algorithm $\mathcal{A}$ crosses the threshold $\eta$, i.e.
\[T_s(\eta) = \inf\{t \in \mathbb{R}_{+}: R(t,y) \geq \eta\}.\]
The performance of the algorithm $\mathcal{A}$ is evaluated in terms of the positive predictive value (PPV), and the true positive rate (TPR) defined as follows
\begin{align}
\mbox{PPV} &= \frac{\mathbb{P}(T_s(\eta) \leq T_{end}|\mathcal{H}_{1})}{\mathbb{P}(T_s(\eta) \leq T_{end}|\mathcal{H}_{o}) + \mathbb{P}(T_s(\eta) \leq T_{end}|\mathcal{H}_{1})},
\label{eqq17} 
\end{align} 
and
\begin{align}
\mbox{TPR} &= \frac{\mathbb{P}(T_s(\eta) \leq T_{end}|\mathcal{H}_{1})}{\mathbb{P}(T_s(\eta) \leq T_{end}|\mathcal{H}_{1}) + \mathbb{P}(T_s(\eta) > T_{end}|\mathcal{H}_{1})},
\label{eqq18} 
\end{align} 
where $T_{end}$ is the time at which observations of the patient's monitored physiological stream stops either because of an ICU admission or discharge (i.e. for a clinically deteriorating patient $T_{end} = (K-\bar{k}+1) \cdot T_1	$).  
		
\subsection{Algorithm}
In this section, we propose an implementation for the algorithm $\mathcal{A}_{off}$ that learns the parameters of the physiological model presented in Section II from a dataset $\mathcal{D}$, and an implementation for the algorithm  $\mathcal{A}_{on}$ which infers the clinical status and computes the risk score for a newly hospitalized patient according to (\ref{eqq15}) and (\ref{eqq16}). The implementation of the algorithms $\mathcal{A}_{off}$ and $\mathcal{A}_{on}$ is confronted with the following challenges:

\begin{enumerate}     
\item The number of patient subtypes $G$ is unknown, and the subtype memberships of the reference patients is not declared in $\mathcal{D}$.
\item The relationship between the admission information $Y$ and the latent subtype $Z$ is unknown and needs to be learned from the data.   
\item The physiological model for the clinically deteriorating patients is non-stationary, and hence, for newly admitted patients, we need to estimate the latent epoch index $\bar{k}$ in real-time in order to synchronize the patient's physiological signal with our model, and properly compute the patient's risk score described by (\ref{eqq15}) and (\ref{eqq16}).
\item The physiological model for the clinically deteriorating patients has many parameters (i.e. $K\left(\frac{D(D+1)}{2}+D+1\right)$ parameters), but the number of clinically deteriorating patients in the dataset $\mathcal{D}$ is relatively small (ICU admission rate is usually less than 10$\%$). \\
\end{enumerate}

In the following, we provide an implementation for the offline algorithm $\mathcal{A}_{off}$ that addresses challenges (1-3), and then we present an implementation for the online algorithm $\mathcal{A}_{on}$ that addresses challenge (4).\\ 
\\
{\bf The offline algorithm $\mathcal{A}_{off}$}\\
\\
The objective of the offline algorithm $\mathcal{A}_{off}$ is to learn from $\mathcal{D}$ the number of subtypes $G$, the parameter set $({\bf \Theta}^{1}_{o},.\,.\,., {\bf \Theta}^{G}_{o},{\bf \Theta}^{1}_{1},.\,.\,., {\bf \Theta}^{G}_{1})$, and the probability of a patient's membership in each subtype given her admission information, i.e. $\mathbb{P}(Z=z|Y=y)$. In the rest of this Section, we use the following notations
\[\Gamma_o = ({\bf \Theta}^{1}_{o},.\,.\,., {\bf \Theta}^{G}_{o}),\] 
\[\Gamma_1 = ({\bf \Theta}^{1}_{1},.\,.\,., {\bf \Theta}^{G}_{1}),\] 
\[\beta_{z}(y) = \mathbb{P}(Z=z|Y=y).\]

Recall from (\ref{eqq15}) that the risk score $R(t,y)$ can be written as 
\begin{align}
R(t,y) = \sum_{z \in \mathcal{Z}} R_{z}(t) \cdot \beta_{z}(y),
\label{eqq19} 
\end{align}
where
\begin{align}
R_{z}(t) = \mathbb{P}\left(V=1\left|\{x_{ij}, t_{ij}\}_{i,j}, {\bf \Theta}^{z}_{o}, {\bf \Theta}^{z}_{1}\right.\right).
\label{eqq19x} 
\end{align} 
The formulation of the risk score $R(t,y)$ in (\ref{eqq19}) explicates the impact of the patient's latent subtype on her risk assessment. The score $R(t,y)$ is a weighted average of the posterior probabilities $R_{z}(t) = \mathbb{P}\left(V=1\left|\{x_{ij}, t_{ij}\}_{i,j}, {\bf \Theta}^{z}_{o}, {\bf \Theta}^{z}_{1}\right.\right)$, i.e. the probabilities of the alternative hypothesis $\mathcal{H}_{1}$ given the evidential physiological data and the latent subtype being $Z=z$, over all possible latent subtypes for the patient. The weight $\beta_{z}(y)$ associated with the term $R_{z}(t)$ corresponds to the probability that the patient with admission information $Y=y$ belongs to subtype $Z=z$. We denote $R_{z}(t)$ as the ``{\it expert for subtype $z$}", whereas the weight $\beta_{z}(y)$ is denoted as the ``{\it responsibility of expert $z$}". Therefore, computing the risk score $R(t,y)$ entails invoking a {\it mixture} of GP experts, and assigning the mixture weights in accordance to the experts' responsibilities determined by $\beta_{z}(y)$.\\  

The algorithm $\mathcal{A}_{off}$ operates in 3 steps. In step 1, we {\it discover the experts}, i.e. we apply the expectation-maximization (EM) algorithm to the dataset $\mathcal{D}_{o}$ in order to estimate the latent patient subtypes and the physiological model parameters for the clinically stable patients. We apply the Bayesian Information Criterion (BIC) for model selection in order to select the number of subtypes $G$. This ensures statistical efficiency in learning the number of subtypes and the model parameters since the physiological model for the clinically stable patients in $\mathcal{D}_{o}$ has only $\frac{D(D+1)}{2}+D+1$ parameters. In step 2, we use a {\it transductive transfer learning approach} to learn the {\it experts' responsibilities} $\beta_{z}(y)$ as a function of the admission information. Finally, in step 3, we use a {\it self-taught} transfer learning approach to learn the parameters of the physiological model for the clinically deteriorating patients through the dataset $\mathcal{D}_{1}$ using the model learned for the clinically stable patients from the dataset $\mathcal{D}_{o}$. In the following, we specify the detailed steps of the algorithm $\mathcal{A}_{off}$.\\     
\\
{\bf \underline{Step 0.} Align the temporal physiological streams in the dataset $\mathcal{D}_{1}$:} Before implementing the 3 steps of the algorithm $\mathcal{A}_{off}$, we need to ensure that the recorded (non-stationary) physiological streams in $\mathcal{D}_{1}$are aligned with respect to a common reference time in order to properly estimate the GP parameters for every epoch $k \in \{1,2,.\,.\,.,K\}$. This is achieved by considering the ICU admission time as a surrogate marker for the latent epoch index $k$. That is, we consider that the samples in the last $T_1$ period of time in every physiological streams to be designated as epoch $K$ (i.e. the last epoch), and then we go backwards in time and label the preceding epochs as $K-1, K-2,$ etc. This procedure is applied to all the physiological streams of the reference patients in $\mathcal{D}_{1}$, and hence all the training physiological streams become aligned in time which allows for a straight-forward epoch-specific parameter estimation.  \\ 
\\
{\bf \underline{Step 1.} Discover the Experts through Clinically Stable Patients:} In this step, we learn both the number of subtypes $G$ (which is also the number of experts), as well as the parameter sets $\Gamma_{o}$. This is accomplished through an iterative approach in which we use the expectation-maximization (EM) algorithm for estimating the parameters in $\Gamma_{o}$ for given values of $G$, and then use the Bayesian information criterion (BIC) to select the number of experts. \\

The detailed implementation of the EM algorithm is given in lines 4-18 in Algorithm 1. The algorithm is executed on the dataset $\mathcal{D}_{o}$ by iterating over the values of $G$, with an initial number of experts $G=1$. For every $M$, we implement the usual E-step and M-step of the EM-algorithm: starting from an initial parametrization $\Gamma_{o}$, in the $p^{th}$ iteration of the EM-algorithm, the auxiliary function $Q(\Gamma_{o};\Gamma_{o}^{p-1})$ is computed as
\[Q(\Gamma_{o};\Gamma_{o}^{p-1}) = \mathbb{E}\left[\mbox{log}\left(\mathbb{P}\left(\left. \mathcal{D}_{o},\{Z^{(n)}\}_{n=1}^{N_o}\right|\Gamma_{o}\right)\right)\left|\mathcal{D}_{o},\Gamma_{o}^{p-1}\right.\right],\] 
where $Z^{(n)}$ is the latent subtype of the $n^{th}$ entry of the dataset $\mathcal{D}_{o}$. The parametrization is updated in the M-step by maximizing $Q(\Gamma_{o};\Gamma_{o}^{p-1})$ with respect to $\Gamma_{o}$ (closed-form expressions are available for the jointly Gaussian data in $\mathcal{D}_{o}$ as per the GP model). The $p^{th}$ iteration is concluded by updating expert $z$'s responsibility towards the $n^{th}$ patient in the dataset $\mathcal{D}_{o}$ as follows
\begin{align}
\beta^{(n)}_{z,p} &= \mathbb{P}\left(Z^{(n)} = z\left|\left\{x^{(n)}_{ij}, t^{(n)}_{ij}\right\}_{i,j}, \Gamma^{p}_{o}\right.\right) \nonumber \\
&= \frac{\pi^{p}_{z} \, f\left(\left.\left\{x^{(n)}_{ij}, t^{(n)}_{ij}\right\}_{i,j}\right|{\bf \Theta}^{p,z}_{o}\right)}{\sum_{z^{'}=1}^{G} \pi^{p}_{z^{'}} \, f\left(\left.\left\{x^{(n)}_{ij}, t^{(n)}_{ij}\right\}_{i,j}\right|{\bf \Theta}^{p,z^{'}}_{o}\right)},
\label{eqq20} 
\end{align}
where $\pi^{p}_{z}$ is the estimate for $\mathbb{P}(Z=z)$ in the $p^{th}$ iteration, and $f(.)$ is the Gaussian distribution function. The term $\beta^{(n)}_{z,p}$ represents the posterior probability of patient $n$'s membership in subtype $z$ given the realization of her physiological data $\left\{x_{ij}, t_{ij}\right\}_{i,j}$. The iterations of the EM-algorithm stop when the claimed responsibilities of the $G$ experts towards the $N_{o}$ reference patients in $\mathcal{D}_{o}$ converges to within a precision parameter $\epsilon$ (line 14).
 
\begin{algorithm}[tb]
   \caption{The Offline Algorithm $\mathcal{A}_{off}$}
   \label{alg:example}
\begin{algorithmic}[1]
   \STATE {\bfseries Input:} Dataset $\mathcal{D}$, precision level $\epsilon$.	 
	 \STATE {\bfseries Implement step 1 (Discover the experts):}
	 \STATE Extract dataset $\mathcal{D}_{o}$ of clinically stable patients with label $v^{(n)}=0$.
	 \STATE Initialize $G=1$ 
   \REPEAT
	 \STATE $p \leftarrow 1$
   \STATE Initialize $\Gamma_{o}^{p} = \{\Theta^{p,z}_{o}\}_{z=1}^{G}$. 
   \REPEAT
   \STATE {\bf E-step:} Compute $Q(\Gamma_{o};\Gamma_{o}^{p-1}).$
   \STATE {\bf M-step:} $({\bf \Theta}^{p}_{o},\{\pi^{p}_{z}\}_{z=1}^{G}) = \mbox{arg} \, \mbox{max}_{\Gamma_{o}} Q(\Gamma_{o};\Gamma_{o}^{p-1}).$
	 \STATE $Q_{G}^{*} \leftarrow \mbox{max}_{\Gamma_{o}} \mbox{max}_{\Gamma_{o}} Q(\Gamma_{o};\Gamma_{o}^{p-1}).$
	 \STATE Update responsibilities using Bayes rule $\beta^{(n)}_{z,p} = \frac{\pi^{p}_{z} \, f(\{x^{(n)}_{ij}, t^{(n)}_{ij}\}_{i,j}|{\bf \Theta}^{p,z}_{o})}{\sum_{z^{'}=1}^{G} \pi^{p}_{z^{'}} \, f(\{x^{(n)}_{ij}, t^{(n)}_{ij}\}_{i,j}|{\bf \Theta}^{p,z^{'}}_{o})}$
   \STATE $p \leftarrow p+1.$
	 \UNTIL{$\frac{1}{N_{o}G}\sum_{i=1}^{N_{o}}\sum_{z=1}^{G}\left|\beta^{(n)}_{z,p}-\beta^{(n)}_{z,p-1}\right| < \epsilon$}
	 \STATE $\Psi_{G} = G\left(\frac{D(D+1)}{2}+D+1\right)$
	 \STATE $B_{G,G-1} \approx \frac{\mbox{exp}\left(Q_{G}^{*}-\frac{1}{2}\Psi_{G}\mbox{log}(N_{o})\right)}{\mbox{exp}\left(Q_{G-1}^{*}-\frac{1}{2}\Psi_{G-1}\mbox{log}(N_{o})\right)}$
	 \STATE $G \leftarrow G+1.$
   \UNTIL{$B_{G,G-1} < \bar{B}$}
	\STATE {\bfseries Implement step 2 (Recruit the experts):}
	\STATE Construct the dataset $\left\{y^{(n)},(\beta^{(n)}_{1},.\,.\,.,\beta^{(n)}_{G})\right\}_{n=1}^{N_o}$.
	\STATE Find linear regression coefficients for $\beta_{z}(y) = [w^z_{1},.\,.\,.,w^z_{S}]^{T}\,y$.
	\STATE {\bfseries Implement step 3 (Self-taught learning):}
	\STATE For every $n \in \mathcal{D}_{1}$ and $z \in \{1,.\,.\,.,G\}$, sample a random variable $c_{n,z} \sim \mbox{Bernoulli}(\beta_{z}(y^{(n)}))$.
	\STATE For every expert $z$, construct a dataset $\mathcal{D}_{1,z} = \{n \in \mathcal{D}_{1}: c_{n,z}=1\}$.
	\STATE Find the MLE estimates of $\Gamma_1$ using the samples in the corresponding datasets $\left\{\mathcal{D}_{1,1},.\,.\,.,\mathcal{D}_{1,G}\right\}$. 
\end{algorithmic}
\end{algorithm}

After each instantiation of the EM-algorithm, we compare the model with $G$ experts to the previous model with $G-1$ experts found in the previous iteration. Comparison is done through the Bayes factor $B_{G,G-1}$ (computed in line 16 via the BIC approximation), which is simply a ratio between Bayesian criteria that trade-off the likelihood of the model being correct with the model complexity (penalty for a model with $G$ experts is given by $\Psi_{G}$ in line 15, such a penalty corresponds to the total number of hyper-parameters in the model with $G$ experts). We stop adding new experts when the Bayes factor $B_{G,G-1}$ drops below a predefined threshold $\bar{B}$.\\ 
\\
{\bf \underline{Step 2.} Recruit the Experts via Transductive Transfer Learning\footnote{Our terminologies with respect to transfer learning paradigms follow those in \cite{pan2010survey}.}:} Having discovered the experts by learning the parameter set $\Gamma_o = ({\bf \Theta}^{1}_{o},.\,.\,., {\bf \Theta}^{G}_{o})$, we need to learn how to associate different experts to the patients based on the initial information we have about them, i.e. the admission features (e.g. transfer status, age, gender, ethnicity, etc). In other words, we aim to learn a mapping rule $\beta_{z}(y): \mathcal{Y} \rightarrow \mathcal{Z}$. The function $\beta_{z}(y)$ reflects the extent to which we rely on the different experts when scoring the risk of a patient with admission information $Y=y$. 

A transductive transfer learning approach is used to learn the function $\beta_{z}(y)$. That is, we use the estimates for the posterior $\beta^{(n)}_{z}$ obtained from step 1 (see line 12 in Algorithm 1) for every patient $n$ in $\mathcal{D}_{o}$, and then we label the dataset $\mathcal{D}_{o}$ with these posteriors, and transfer these labels to the domain of admission features, thereby constructing a dataset of the form $\left\{y^{(n)},(\beta^{(n)}_{1},.\,.\,.,\beta^{(n)}_{G})\right\}_{n=1}^{N_o}$. We use a linear regression analysis to fit the function $\beta^{(n)}_{z}$ (see lines 20-21 in Algorithm 1).\\
\\
{\bf \underline{Step 3.} Perform a Self-taught Discovery for the Experts of Clinically Deteriorating Patients:} The knowledge of the parameter set $\Gamma_1 = ({\bf \Theta}^{1}_{1},.\,.\,., {\bf \Theta}^{G}_{1})$ needs to be gained from the dataset $\mathcal{D}_{1}$. We use a self-taught transfer learning approach to transfer the knowledge obtained using unsupervised learning from the dataset $\mathcal{D}_{o}$, i.e. the domain of stable patients, to ``label" the dataset $\mathcal{D}_{1}$ and learn the set of experts associated with the clinically acute patients \cite{pan2010survey},\cite{raina2007self}. 

\begin{figure*}[t!]
    \centering
    \includegraphics[width=5.5 in]{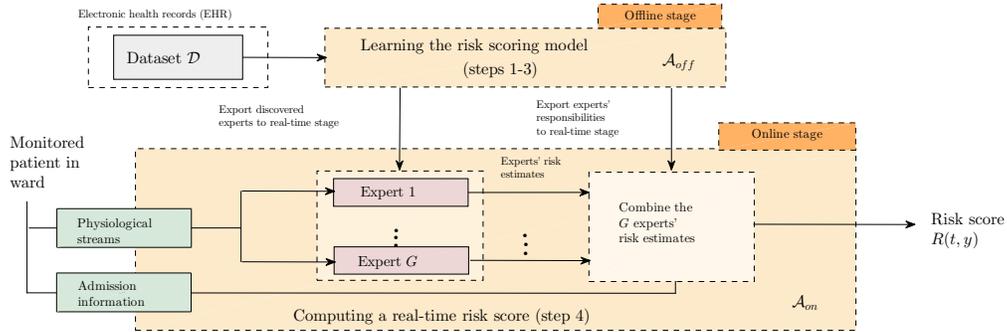}     
    \caption{Block diagram for the proposed risk scoring algorithm.}
		\label{bckdg}
\end{figure*}

Self-taught learning is implemented by exporting the number of experts $G$ that we estimated from $\mathcal{D}_{o}$ directly to the population of patients in $\mathcal{D}_{1}$, picking a subset of patients in $\mathcal{D}_{1}$ to estimate the parameter set ${\bf \Theta}^{z}_{1}$ of expert $z$ by sampling patients from $\mathcal{D}_{1}$ using their responsibility vectors (line 23 in Algorithm 1). \\
\\
{\bf The online algorithm $\mathcal{A}_{on}$}\\
\\
An aggregate risk score for every patient with admission information $Y=y$ is obtained by weighting the opinions of the $G$ experts with their responsibilities $\{\beta_{z}(y)\}_{z=1}^{G}$. The risk score for a newly hospitalized patient $i$ with admission information $Y=y$ at time $t$ is then given by 
\[R(t,y) = \sum_{z=1}^{G}\frac{\beta_{z}(y)}{\sum_{z^{'}=1}^{G}\beta_{z^{'}}(y)}\,R_{z}(t).\]
Note that computing $R_{z}(t)$ is not possible unless we know the latent epoch index $\bar{k}$ for the monitored patient. Since $\bar{k}$ is a hidden variable, we estimate $\bar{k}$ and evaluate $R_{z}(t)$ by averaging over its posterior distribution, i.e. 
\begin{align} 
R_{z}(t) &= \mathbb{E}_{\bar{k}}\left[\mathbb{P}(V=1|\{x_{ij}, t_{ij}\leq t\}_{ij}, \bar{k}, \Gamma_o, \Gamma_1)\right]\nonumber \\
&= \sum_{1 \leq \bar{k} \leq K} \mathbb{P}(V=1|\{x_{ij}, t_{ij}\leq t\}_{ij}, \bar{k}, \Gamma_o, \Gamma_1) \, \times \nonumber \\
& \,\,\,\,\, \mathbb{P}(\bar{k}|\{x_{ij}, t_{ij}\leq t\}_{ij}, \Gamma_1), \nonumber \\
\end{align}
where $\mathbb{P}(V=1|\{x_{ij}, t_{ij}\leq t\}_{ij}, \bar{k}, \Gamma_o, \Gamma_1)$ is evaluated via Bayes rule as clarified in (\ref{eqq16}). Hence, the online algorithm $\mathcal{A}_{on}$ continuously estimates the latent epoch index $\bar{k}$ as more physiological data is gathered, and synchronized the monitored physiological stream with the learned (non-stationary) GP model.\\

Algorithm 2 shows the a pseudo-code for the operations implemented in the real-time stage. Fig. \ref{bckdg} illustrate a block diagram with all the steps of the $\mathcal{A}_{off}$ and $\mathcal{A}_{on}$ algorithms. 
\begin{algorithm}[tb]
   \caption{The Online Algorithm $\mathcal{A}_{on}$}
   \label{alg:example}
\begin{algorithmic}[1]
   \STATE {\bfseries Input:} Physiological measurements $\left\{x_{ij}, t_{ij}\right\}_{i,j}$, admission features $Y = y$, a set of experts' parameters $\Gamma_o$ and $\Gamma_1$.
	 \STATE Estimate the experts' responsibilities $\beta_{z}(y)$.
	 \STATE Compute the posterior epoch index distribution $\mathbb{P}(\bar{k}|\{x_{ij}, t_{ij}\leq t\}_{ij},\Gamma_1)$.
	 \STATE For every expert $z$, compute the risk score 
	\[R_{z}(t) = \sum_{1 \leq \bar{k} \leq K} \mathbb{P}(V=1|\{x_{ij}, t_{ij}\leq t\}_{ij}, \bar{k}, \Gamma_o, \Gamma_1) \, \times\] 
	\[\mathbb{P}(\bar{k}|\{x_{ij}, t_{ij}\leq t\}_{ij}, \Gamma_1).\]
   \STATE Compute the final risk score as a mixture of the individual experts' risk assessments weighted by their individual responsibilities toward the monitored patient  
	\[R(t,y) = \sum_{z=1}^{G}\frac{\beta_{z}(y)}{\sum_{z^{'}=1}^{G}\beta_{z^{'}}(y)}\,R_{z}(t).\] 
\end{algorithmic}
\end{algorithm}

\section{Experiments and Results}
In order to evaluate its clinical utility, we have applied our risk scoring algorithm to a cohort of patients who were recently admitted to a general medicine floor in the Ronald Reagan UCLA medical center. We start by describing the patient cohort in the following subsection, and then we present the results of our experiments illustrating the performance of the proposed risk scoring scheme.

\begin{table*}[t]
  \centering
  \begin{tabular}{c|c|c}
    \hline
		\hline
    Vital signs & Lab tests & Admission information \\
    \hline
		&  &  \\
		Diastolic blood pressure & Glucose & Transfer Status \\
    Eye opening & Urea Nitrogen  & Gender \\
		Glasgow coma scale score & White blood cell count & Age \\
		Heart rate & Creatinine & Stem cell transplant \\
		Respiratory rate  & Hemoglobin & Floor ID \\
		Temperature & Platelet Count & ICD-9 codes \\
		$O_2$ Device Assistance & Potassium & Race \\
		$O_2$ Saturation & Sodium & Ethnicity \\
		Best motor response & Total $CO_2$ &  \\
		Best verbal response & Chloride &  \\
		Systolic blood pressure &  &  \\
		 &  &  \\
    \hline
		\hline
  \end{tabular}
	\captionsetup{font= small}
  \caption{Physiological data and admission information associated with the patient cohort under study.}
\end{table*}

\subsection{Data Description}
\begin{figure*}[t!]
    \centering
    \includegraphics[width=6.5 in]{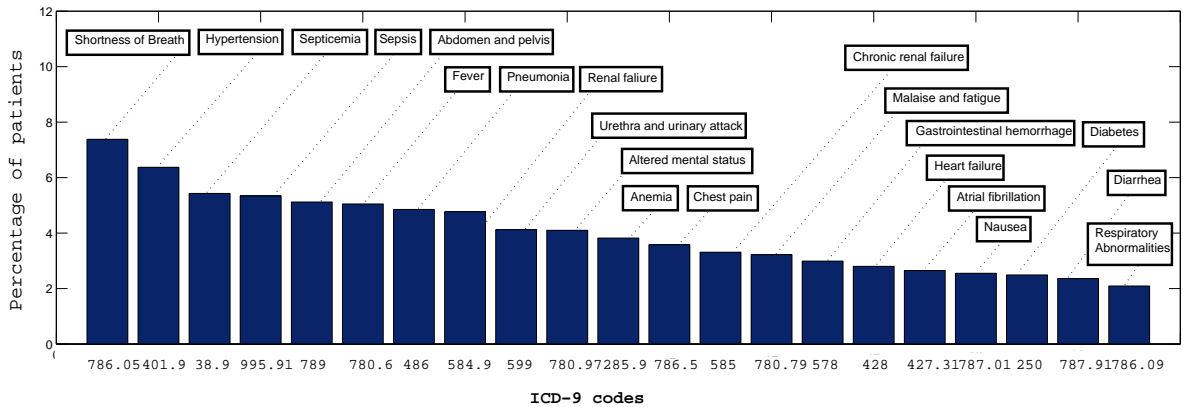}  
	  \captionsetup{font= small}
    \caption{Distribution of the ICD-9 codes in the patient cohort.}
		\label{provervw}
\end{figure*}

Experiments were conducted on a cohort of 6,321 patients who were hospitalized in a general medicine floor during the period between March $3^{rd}$ 2013, to February $4^{rd}$ 2016. The patients' population is heterogeneous with a wide variety of diagnoses and ICD-9 codes. The distribution of the ICD-9 codes associated with the patients in the cohort is illustrated in Fig. \ref{provervw}. The cohort included patients who were not on immunosuppression and others who were on immunosuppression, including patients that have received solid organ transplantation. In addition, there were some patients that had diagnoses of leukemia or lymphoma. Some of these patients received stem cell transplantation as part of their treatment. Because these patients receive chemotherapy to significantly ablate their immune system prior to stem cell transplantation, they are at an increased risk of clinical deterioration. Of the 6,321 patients (the dataset $\mathcal{D}$), 524 patients experienced clinical deterioration and were admitted to the ICU (the dataset $\mathcal{D}_1$), and 5,788 patients were discharged home (the dataset $\mathcal{D}_o$). Thus, the ICU admission rate is 8.30$\%$.

Patients in the dataset $\mathcal{D}$ were monitored for 11 vital signs (e.g. $O_{2}$ saturation, heart rate, systolic blood pressure, etc) and 10 lab tests (e.g. Glucose, white blood cell count, etc). Hence, the dimension of the physiological stream for every patient is $D=21$. Table II lists all the vital signs and lab tests included in the experiment, in addition to the set of admission information $Y$ that are used for personalizing the computed risk scores. The sampling rate for the physiological streams $\{x_{ij}, t_{ij}\}_{i,j}$ ranges from 1 hour to 4 hours, and the length of hospital stay for the patients ranged from 2 to 2,762 hours. Correlated feature selection (CFS) was used to select the physiological streams that are relevant to predicting the endpoint outcomes (i.e. ICU admission); the CFS algorithm selected 7 vital signs (Diastolic blood pressure, eye opening, Glasgow coma scale score, heart rate, temperature, $O_2$ device assistance and $O_2$ saturation), and 3 lab tests (Glucose, Urea Nitrogen and white blood cell count). The CFS algorithm is optimized for every competing algorithm involved in the comparisons presented in this Section. 

Throughout the experiments conducted in this Section, the training and testing datasets are constructed as follows. The training set comprises 5,130 patients who were admitted to the ward in the period between March 2013 and July 2015. Among those patients, the ICU admission rate was $8.34 \%$. The algorithms are trained via this dataset, and then tested on a separate dataset that comprises the remaining 1,191 patients who were admitted to the ward in the period between July 2015 and April 2016. Thus, all the algorithms involved in the experiments are tested on the most recently hospitalized patients in the cohort. In the next Subsection, we highlight the patients' subtypes discovered by our algorithm, and the consequent clinical insights associated with these discoveries.       

\subsection{Subtype Discovery}
\begin{figure}[t!]
    \centering
    \includegraphics[width=3.5 in]{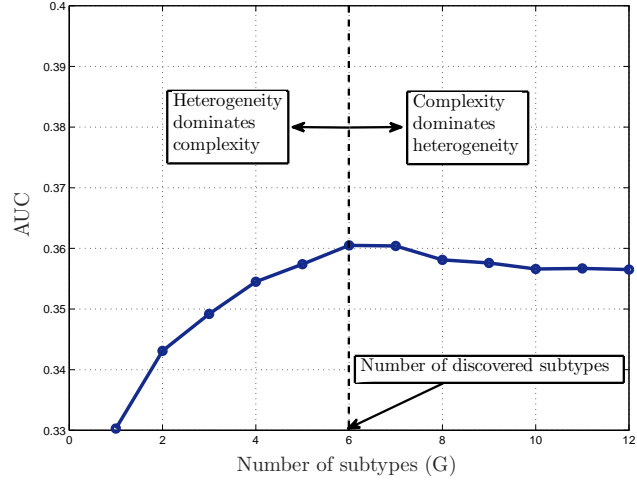}    
		\captionsetup{font= small}
    \caption{AUC performance for different number of subtypes $G$.}
		\label{Fiqsubtype}
\end{figure}

When running the risk scoring algorithm on the 5,130 patients in the testing set, the algorithm was able to discover 6 patient subtypes ($G=6$), and train the corresponding GP experts. Figure \ref{Fiqsubtype} demonstrates the area under curve (AUC) performance of the proposed algorithm versus the number of subtypes $G$. For $G<6$, the gain attained by capturing the heterogeneity of the patients' population dominates the losses endured due to the increased model complexity. For $G>6$, adding more subtypes increases the complexity of the risk model without capturing further heterogeneity, and hence the performance degrades. Setting the number of subtypes as $G=6$ experts is optimal given the size of the dataset $\mathcal{D}$; the offline algorithm $\mathcal{A}_{off}$ stops after computing the Bayes factor $B_{6,5}$ (see line 18 in Algorithm 1). If the algorithm is to be applied to a larger dataset drawn from the same population, the peak in Figure \ref{Fiqsubtype} would shift to the right, i.e. more patient subtypes would be discovered leading to a more granular risk model.\\  

Having discovered the latent patient subtypes, we investigate how the hospital admission features $Y$ are associated to the patients' subtypes, i.e. we are interested in understanding {\it which} of the admission features are most representative of the latent patient subtypes. Table III lists the admission features ranked by their ``importance" in deciding the responsibilities of the 6 experts corresponding to the 6 subtypes. The importance, or relevance, of an admission feature is quantified by the weight of that feature $(w_{1},.\,.\,.,w_{S})$ in the learned linear regression function $\beta_z(y)$ (see line 21 in Algorithm 1). \\

As shown in Table III, stem cell transplant turned out to be the feature that is most relevant to the assignment of responsibilities among experts. This is consistent with domain knowledge: patients receiving stem cell transplantation are at a higher risk of clinical deterioration due to their severely compromised immune systems, thus it is extremely important to understand their physiological state \cite{hayani2011impact}. This is borne out in Table III as stem cell transplantation status has the largest contribution in selecting the suitable experts. We note that, in Ronald Reagan medical center, patients with leukemia and lymphoma are often taken care of on the same floor as the general medicine population. This then demonstrates the point that it is crucial to utilize information about the heterogeneity of patients to improve their personalized medical care. Table III also shows that the floor ID is relevant to the patient's latent subtype, which follows from the fact that different floors are likely to accommodate patients with different diagnoses.\\

Surprisingly, gender turned out to be the third most relevant feature for expert assignments. This means that vital signs and lab tests for males and females should not be interpreted in the same way when scoring the risk of clinical deterioration, i.e. different GP experts needs to handle different genders (recall the demonstration in Fig. \ref{Fiq3}). The fact that the transfer status of a patient is an important admission factor (ranked fourth in the list) is consistent with prior studies that demonstrate that patients transferred from outside facilities have a higher acuity with increased mortality \cite{rincon2011association}.

\begin{table}[t]
  \centering
  \begin{tabular}{c|c|c}
    \hline
		\hline
    Rank & Admission feature & Regression coefficient \\
    \hline
		&  &  \\
		1 & Stem cell transplant & 0.1091 \\
    2 & Floor ID & 0.0962  \\
		3 & Gender & 0.0828 \\
		4 & Transfer status & 0.0827 \\
		5 & ICD-9 code & 0.0358  \\
		6 & Age & 0.0109 \\
		 &  &  \\
    \hline
		\hline
  \end{tabular}
	\captionsetup{font= small}
  \caption{Relevance of the patients' admission features to the latent subtype memberships.}
\end{table}

\subsection{Prognosis and Early Warning Performance}

\begin{table*}[t]
  \centering
  \begin{tabular}{p{0.75cm}|p{2.75cm}|p{0.5cm}|p{1.25cm}|p{1cm}|p{0.5cm}|p{1cm}|p{1cm}|p{1cm}|p{1cm}}
    \hline
		\hline
    TPR & Proposed score ($G=6$)  & LR$^{*}$ & Logit. R.$^{*}$ & LASSO  & RF$^{*}$ & MEWS & SOFA & APACHE & Rothman \\
    \hline
		  &  &  &  &  &  &  &  &  &  \\
		40$\%$ & \centering 1.76 & \centering 2.58 & \centering 2.3	& \centering 2.3 &	\centering 3.31 & \centering 5.9 & \centering 7.26 & \centering 6.41 &  3.98  \\
    50$\%$ & \centering 2.16 & \centering 4.46 & \centering 3.95 & \centering 3.44 & \centering 4.62 & \centering 7.13 & \centering 7.77 & \centering 7.13 &  4.56  \\
    60$\%$ & \centering 2.44 & \centering 5.13 & \centering 4.99 & \centering 4.95 & \centering 5.45 & \centering 7.06 & \centering 7.06 & \centering 7.77 &  5.62  \\
    70$\%$ & \centering 3.15 & \centering 6.09 & \centering 6.25 & \centering 6.09 & \centering 6.41 & \centering 8.8 & \centering 8.52 & \centering 8.62 &  6.35  \\
    80$\%$ & \centering 4.81 & \centering 6.63 & \centering 7.2 & \centering 7.2 & \centering 6.94 & \centering 9.31 & \centering 9.31 & \centering 9.75 & 7.33  \\
		  &  &  &  &  &  &  &  &  &  \\
    \hline
		\hline
  \end{tabular}
	\captionsetup{font= small}
  \caption{Number of false alarms per one true alarm ($^{*}$ LR = Linear regression, Logit. R. = Logistic regression, and RF = Random forest).}
\end{table*}

\begin{figure}[t!]
    \centering
    \includegraphics[width=3.5 in]{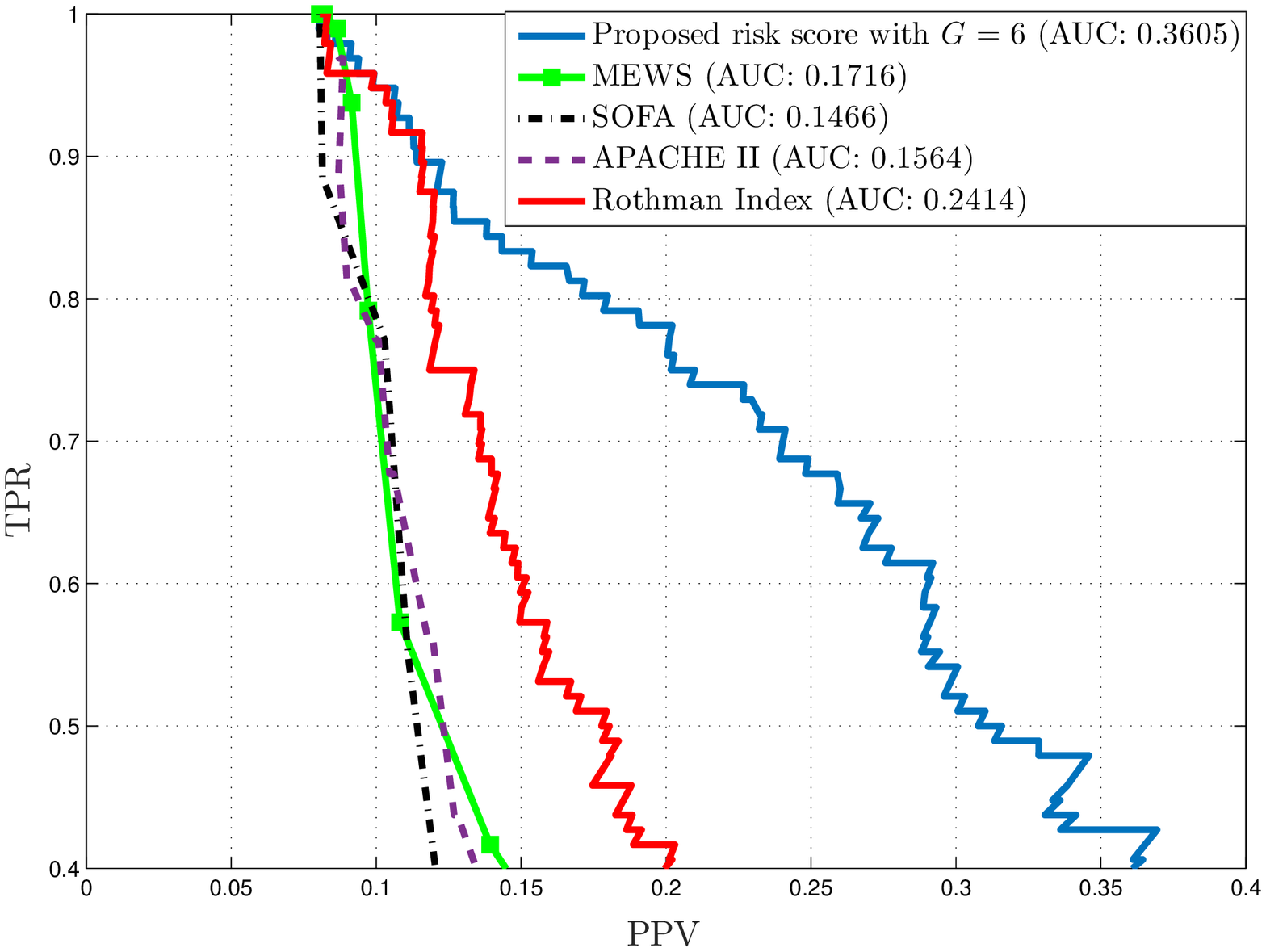}  
	  \captionsetup{font= small}
    \caption{TPR and PPV performance comparisons (ROC curve) with respect to state-of-the-art risk scores.}
		\label{pr1}
\end{figure}
\begin{figure}[t!]
    \centering
    \includegraphics[width=3.5 in]{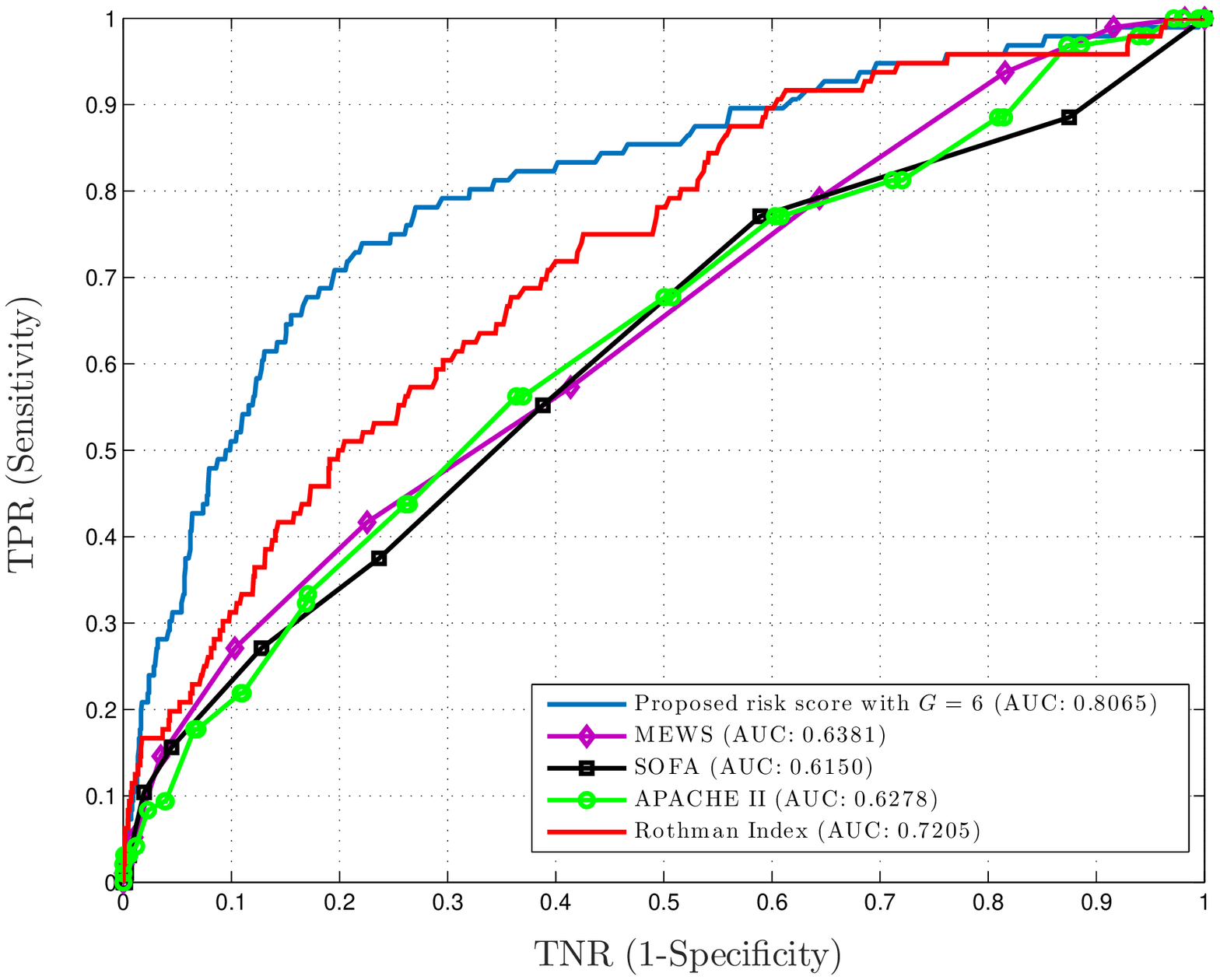}
	  \captionsetup{font= small}
    \caption{TPR and TNR performance comparisons (ROC curve) with respect to state-of-the-art risk scores.}
		\label{pr2}
\end{figure}
\begin{figure}[t!]
    \centering
    \includegraphics[width=3.5 in]{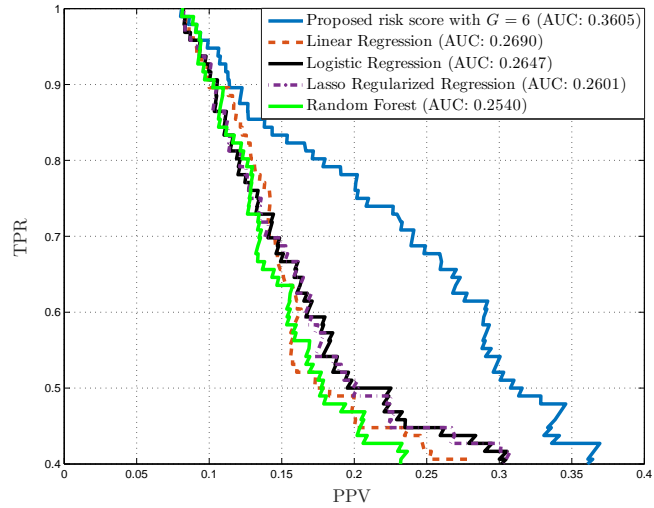}  
		\captionsetup{font= small}
    \caption{TPR and PPV performance comparisons (ROC curve) with respect to state-of-the-art machine learning techniques.}
		\label{pr3}
\end{figure}

\begin{figure*}[t!]
    \centering
    \includegraphics[width=6.5 in]{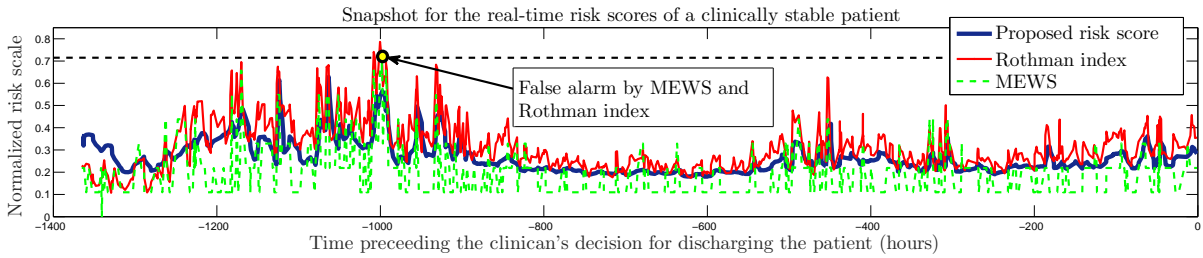}     
    \caption{A sample path for the risk assessment of a clinically stable patient in the testing dataset.}
		\label{pr4}
\end{figure*}

\begin{figure}[t!]
 \centering 
    \includegraphics[width=3.5 in]{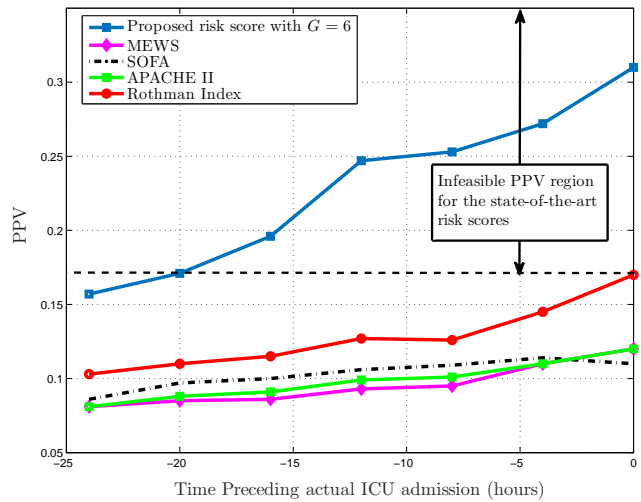}
	  \captionsetup{font= small}
    \caption{Timeliness of the proposed risk score.}
		\label{pr5}
\end{figure}
\begin{figure}[t!]
    \centering
    \includegraphics[width=3.5 in]{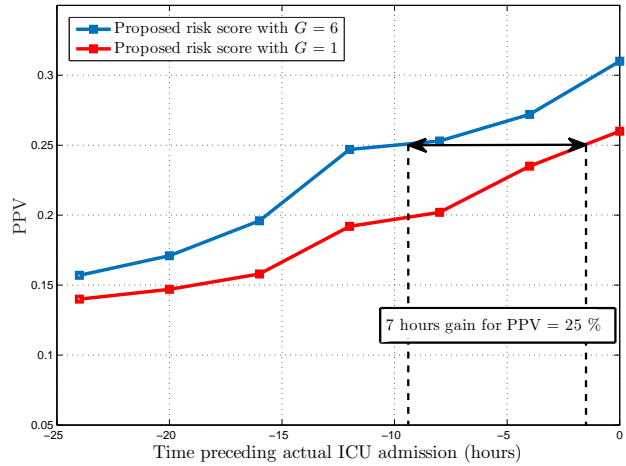}  
		\captionsetup{font= small}
    \caption{Impact of personalization on the timeliness of the ICU alarms.}
		\label{pr6}
\end{figure}
\begin{figure}[t!]
    \centering
    \includegraphics[width=3.5 in]{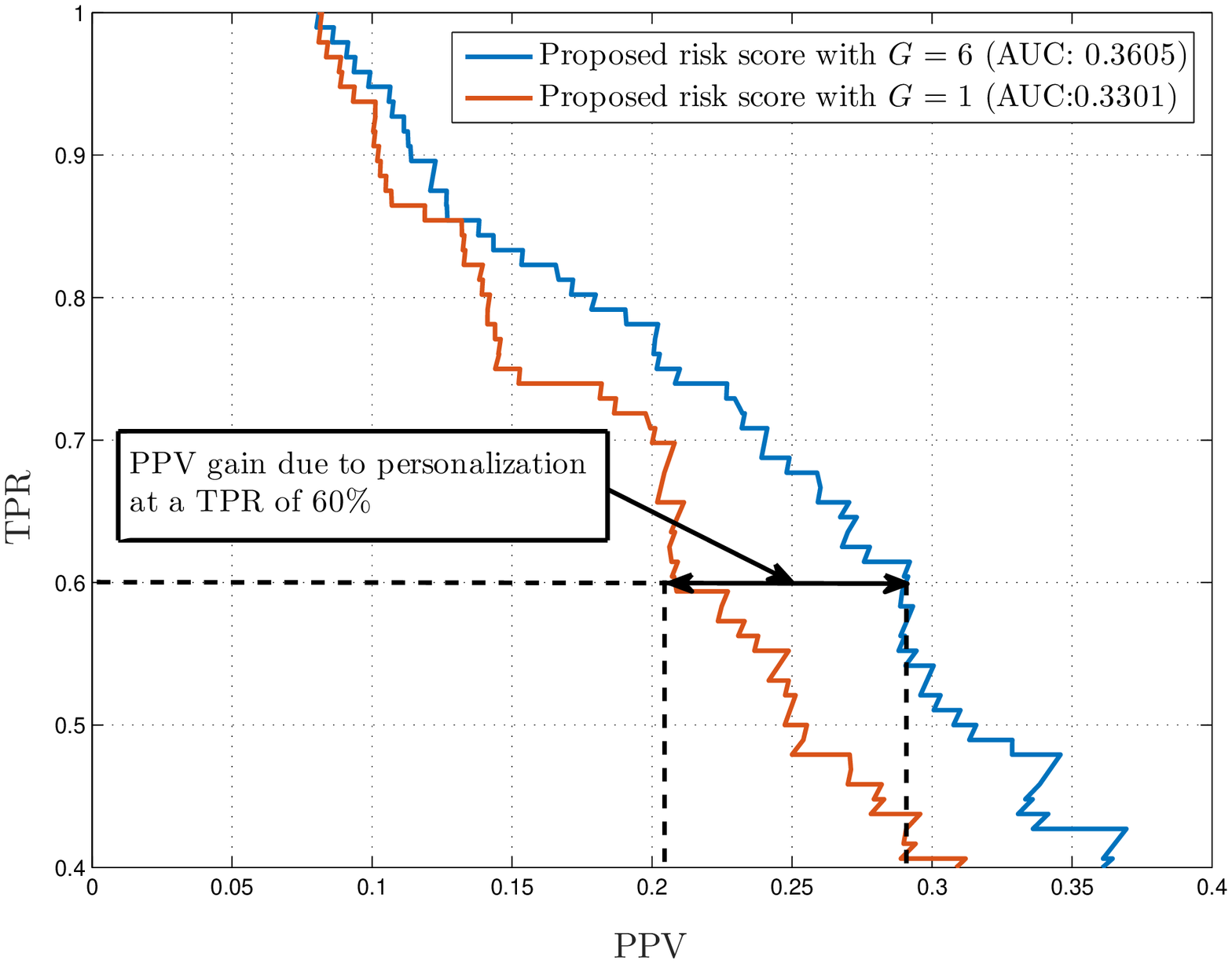}  
	  \captionsetup{font= small}
    \caption{Impact of personalization on the ROC curve.}
		\label{pr7}
\end{figure}

We validated the utility of the proposed risk scoring model by constructing an EWS that issues alarms for ICU admission based on the real-time risk score (i.e. ICU alarms are issued whenever the risk score $R(t,y)$ crosses a threshold $\eta$), and evaluating the performance of the EWS in terms of the PPV and the TPR as defined in (\ref{eqq17}) and (\ref{eqq18}). The accuracy of the proposed risk model is compared with that of the state-of-the-art risk scores (Rothman, MEWS, APACHE and SOFA) by evaluating the Receiver Operating Characteristics (ROC) curves in Fig \ref{pr1}. The implementation of the MEWS and Rothman indexes followed their standard methodologies in \cite{subbe2001validation} and \cite{rothman2013development}, whereas the implementations of SOFA and APACHE followed \cite{yu2014comparison}. \\ 

As shown in Fig. \ref{pr1}, the proposed risk model with $G=6$ subtypes consistently outperforms all the other risk scores for any setting of the TPR and PPV. The proposed score offers gains of $12\%$ with respect to the (most competitive) Rothman score ($p$-value $<$ 0.01). This promising result shows the prognostic value of replacing the currently deployed scores in wards with scores that captures the patients' heterogeneity, considers the temporal aspects of the physiological data, and accounts for the correlations among different physiological streams. The same comparison is carried out in Fig. \ref{pr2}, but in terms of the TPR and the true negative rate (TNR) performances, and it can be seen that the AUC of the proposed score (0.806) outperforms that of the Rothman index (0.72) and all other risk scoring methods. Moreover, as shown in Fig. \ref{pr3}, the proposed risk score also outperforms state-of-the-art machine learning techniques (logistic regression, linear regression, random forest, and LASSO); it provides an AUC gain of around 10$\%$ with respect to these techniques ($p$-value $<$ 0.01). \\  

It is important to note that the proposed risk score significantly reduces the false alarm rates as compared to the state-of-the-art risk scores. This can be seen for the numerical values in Table IV and is also reflected in the TPR/PPV performance comparison in Fig. \ref{pr1}, where we can see that for any fixed TPR, the proposed risk score achieves a much higher PPV than the Rothman index, e.g. at a TPR of 60$\%$, the proposed score achieves a PPV of 30$\%$, which is double of that achieved by the Rothman index (15$\%$). This significant reduction in the false alarm rate can be attributed to the fact that the proposed algorithm computes a risk score based on a trajectory of measurements rather than instantaneous ones. Fig. \ref{pr7} illustrates this effect by depicting a realization for the risk scores' trajectory of a clinically stable patient in the testing dataset. We can see that the MEWS and Rothman indexes exhibit drastic fluctuations over time as they only consider the most recent vital signs and lab tests, which makes them easily triggered by instantaneous measurements or transient phenomena. Our score offers a smoother trajectory that is more resilient to false alarms since it computes a posterior probability that is conditioned on the entire physiological history.\\ 

Reductions in the false alarm rates are further demonstrated in Table IV, where we specify the number of false alarms per one true alarm for both the proposed risk score and the state-of-art scores at different settings of the TPR. At a TPR of 50$\%$, our risk score leads to only 2.16 false alarms for every 1 true alarm, whereas the Rothman index lead to 4.56 false alarms per true alarm, i.e. the rate of the false alarms caused by the Rothman index is more than double of that caused by the proposed algorithm. Thus, our risk score can ensure more confidence in its issued ICU alarms, which would mitigate alarm fatigue and enhance a hospital's resource utilization \cite{cvach2012monitor, mainresult2}. Table IV shows that our risk score offers a consistently lower false alarm rate compared to all other risk scores and machine learning algorithms for all settings of the TPR. \\

Fig. \ref{pr5} illustrates the trade-off between the timeliness of the ICU alarm and its accuracy for a fixed TPR of 50$\%$ (the achieved gains hold for any setting of the TPR). We can see that the proposed risk score consistently outperforms all the other scores in terms of the timeliness of its ICU alarms for all the PPV settings. For instance, for a PPV greater than 25$\%$, our score offers a 12-hour earlier predictions with respect to the actual physician-determined ICU admission event. This level of timeliness is not feasible for any of the other risk scores. Combining the results shown in Fig. \ref{pr5} and Table IV, one can see that the proposed risk score is able to both warn the clinician earlier and provide a more confident signal as compared to the state-of-the-art risk scores, thus providing the ward staff with a safety net for patient care by giving them sufficient time to intervene in order to prevent clinical deterioration. \\

The value of personalization is depicted in Fig. \ref{pr6} and Fig. \ref{pr7}, where we plot the ROC and timeliness curves for our algorithm once with one subtype (i.e. $G=1$ and no personalization is taken into account), and once with $G=6$ subtypes. If we were to take $G=1$, our model would prompt ICU alarms that warns the clinicians 5 hours earlier than the physicians' determination. When we take $G=6$, our model prompts ICU alarms 12 hours earlier than physician determination. Thus, even the unpersonalized version of our model is significantly quicker than the physician determination, but is sluggish in comparison to the personalized one. A similar gain is attained due to personalization in terms of the PPV. As shown in Fig. \ref{pr7}, personalization leads to a 10$\%$ higher PPV at a TPR of 60$\%$ as compared to a non-personalized version of our model.   

\section{Conclusion}
In this paper, we have developed a personalized risk scoring algorithm for critically ill patients in wards that allows transferring deteriorating patients to the ICU in a timely manner. The algorithm learns a granular risk scoring model that is tailored to the individual patient's traits by modeling the patient's physiological processes via a mixture of multitask Gaussian Processes, the weights of which are determined by the patient's baseline admission information and the latent subtypes discovered from the training data. We have demonstrated the utility of the proposed risk scoring algorithm through a set of experiments conducted on a heterogeneous cohort of 6,321 critically ill patients who were recently admitted to Ronald Reagan UCLA medical center. The experiments have shown that the proposed risk score significantly outperforms the currently deployed risk scores, such as the Rothman index, MEWS, APACHE and SOFA scores, in terms of timeliness, true positive rate, and positive predictive value. The results suggest the possibility of reducing the annual sub-acute care mortality rates significantly by applying the concepts of precision medicine. 

\bibliography{ICU_ref}
\bibliographystyle{IEEEtran}

\end{document}